\documentclass[10pt,twocolumn,letterpaper]{article}

\usepackage{cvpr}              % To produce the CAMERA-READY version
\usepackage[dvipsnames]{xcolor}

\usepackage{verbatim}
\usepackage{multirow}
\usepackage{multicol}
\usepackage{float}
\usepackage{stfloats}

\usepackage[ruled,vlined]{algorithm2e}
\usepackage{algpseudocode}

\definecolor{cvprblue}{rgb}{0.21,0.49,0.74}
\usepackage[pagebackref,breaklinks,colorlinks,citecolor=cvprblue]{hyperref}
\usepackage{sistyle}
\SIthousandsep{,}
\def\paperID{2753} % *** Enter the Paper ID here
\def\confName{CVPR}
\def\confYear{2024}

\title{Rethink Predicting the Optical Flow with the Kinetics Perspective}
\author{Yuhao Cheng \hspace{10pt}Siru Zhang \hspace{10pt}Yiqiang Yan\\
Lenovo Research\\
{\tt\small yuhao.cheng@outlook.com}
}

\begin{document}
\maketitle
\begin{abstract}
    %Background, optional or in one sentence
    Optical flow estimation is one of the fundamental tasks in low-level computer vision, which describes the pixel-wise displacement and can be used in many other tasks. From the apparent aspect, the optical flow can be viewed as the correlation between the pixels in consecutive frames, so continuously refining the correlation volume can achieve an outstanding performance. However, it will make the method have a catastrophic computational complexity. Not only that, the error caused by the occlusion regions of the successive frames will be amplified through the inaccurate warp operation. These challenges can not be solved only from the apparent view, so this paper rethinks the optical flow estimation from the kinetics viewpoint.
    We propose a method combining the apparent and kinetics information from this motivation. The proposed method directly predicts the optical flow from the feature extracted from images instead of building the correlation volume, which will improve the efficiency of the whole network. Meanwhile, the proposed method involves a new differentiable warp operation that simultaneously considers the warping and occlusion. Moreover, the proposed method blends the kinetics feature with the apparent feature through the novel self-supervised loss function.
    Furthermore, comprehensive experiments and ablation studies prove that the proposed novel insight into how to predict the optical flow can achieve the better performance of the state-of-the-art methods, and in some metrics, the proposed method outperforms the correlation-based method, especially in situations containing occlusion and fast moving. The code will be public\footnote[1]{\url{https://github.com/YuhaoCheng/OpticalFlow_Kinetic}}.
\end{abstract}    
\vspace{-1.5em}
\section{Introduction}
\label{sec:intro}
\vspace{-0.5em}
\begin{figure}[ht]
    \centering
    \includegraphics[width=0.8\linewidth]{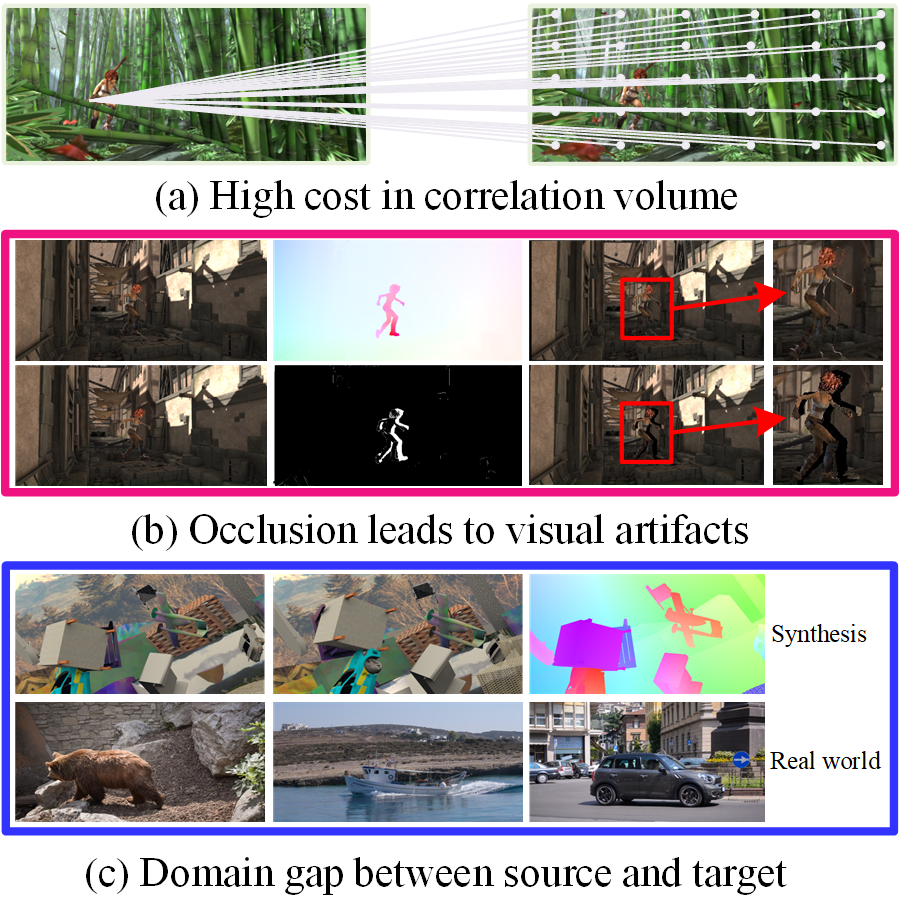}
    \vspace{-1em}
    \caption{Main challenges (better in color). (a) Dense correlation volume represents pixel-wise correspondences between frames. Estimating optical flow based on the dense correlation will bring a high cost both in memory and computation. (b) Occlusion leads to distinct visual artifacts with the warping operation. The first column is two consecutive frames. The upper image in the 2nd column is the optical flow, and the lower is the occlusion map. From the 3rd column, we can know that warping without the occlusion map will cause some unreal visual artifacts(top), and the occlusion map will avoid these(bottom). (c) The upper part is the synthetic data used for training, while the lower part is the situation where the trained models are used. Obviously, these two domains are severely different in texture, however, they have the same characteristics in the kinetics.}
    \label{fig:1_vis_challenges}
    \vspace{-2em}
\end{figure}
Optical flow is the pattern of apparent motion of objects, surfaces, and edges in a visual scene caused by the relative motion between an observer and a scene.
Estimating the optical flow is one of the essential tasks in computer vision. The accurate optical flow will improve the performance of many other computer vision tasks, e.g., video compression, video frame interpolation, action recognition, scene understanding, etc. 
However, accurately estimating the optical flow with low computational requirements is still one of the challenging tasks in the low-level computer vision area. Before the bloom of the learning-based methods, researchers have explored various traditional methods such as ~\cite{brox2010large,weinzaepfel2013deepflow,revaud2015epicflow,bailer2015flow,hur2017mirrorflow} by using the motion priors to predict the optical flow.
With the development of the learning-based method, the performance of optical flow estimation has been improved largely. Nevertheless, there still are some challenges impeding the wide use of learning-based optical flow methods, showing in \cref{fig:1_vis_challenges}, such as high-computation cost, occlusion between objects, the different data distribution between source and target domain, etc. 
In order to alleviate these challenges, we propose a new view to rethink how to predict the motion information. 

Firstly, classical methods utilize the correlation volume to represent the relation between features extracted from images. However, it is high computational complexity and large memory consumption for prevailing dense 4D correlation volume, which increases the overhead of the optical flow estimation methods that already require many computational resources. 
The previous widely used method to estimate the optical flow is using the correlation volume to compute the motion information, as shown in the \cref{fig:1_vis_challenges}(a). Although it can achieve a great performance, the high computation cost makes it difficult to use. 
So, in our method, in order to avoid iterative searching in the correlation volume, we presume that the accurate optical flow can be achieved with discriminative features even without the correlation volume. We introduce a feature motion decoder to predict optical flow directly from the feature domain, which reduces the consumption of computational resources.

Secondly, as shown in~\cref{fig:1_vis_challenges}(b), estimating optical flow in the occluded areas is a perpetual challenge. Pixels visible in one frame and invisible in another will introduce occlusion regions. And when warping in the occluded areas, it will generate distinct visual artifacts, so-called ghost effects, which increase the prediction errors during optical flow estimation. So, in order to address the significant visual artifacts caused by the warping operation in the occluded areas, we propose a novel learnable warping operation.
And the key insight of this module is to combine predicting occlusion areas and warping based on predicted occlusion maps for unified optimization.

Thirdly, the gap between the source and target domain in optical flow is another awkward challenge, as shown in~\cref{fig:1_vis_challenges}(c). Labeling accurate optical flow is so hard that researchers think using the synthesis data to alleviate the severe lack of the ground truth. To some extent, limited real-world optical flow datasets limit the performance to advance further as data is one of the key components to promote learning-based methods. 
So, this paper proposes a self-supervised learning method to use the inherent properties of motion to utilize the unlabeled data.

In conclusion, the major contributions of this paper can be summarized as follows:
\begin{itemize}
    \item Propose a novel motion estimation method using the self-supervised manner with an innovative loss function to leverage the intrinsic feature of kinetics.
    \item Propose a network to directly estimate the optical flow in feature instead of using the correlation volume.
    \item Propose a novel warping module, WarpNet, for more accurate features and images warped in occluded regions based on predicted occlusion maps, which is simultaneously optimized with the whole network.
\end{itemize}

% \vspace{-1em}
\section{Related Work}
\label{sec:related_work}
\subsection{Optical Flow Estimation}
Optical flow is the basic task in the low-level computer vision, so variational methods provide methods to estimate the optical flow such as~\cite{hosni2012fast,Dosovitskiy_FlowNet,ilg2017flownet,sun2018pwc,teed2020raft,jiang2021learning_optical_flow}. \cite{hosni2012fast} firstly proposed using the correlation volume, which is helpful in building pixel-wise correspondences between frames, to predict the optical flow instead of designing different descriptors and searching in the field. However, this kind of method is so resource-consuming that many researchers propose some improvements.
Former methods~\cite{Dosovitskiy_FlowNet,ilg2017flownet,sun2018pwc} construct a local correlation volume only considering the neighbor pixels for matching to save memory, which tends to ignore the large displacements. Some methods~\cite{brox2004highacc,ranjan2017optical} use a coarse-to-fine strategy and build a 4D correlation volume at the coarse resolution to reduce the memory and computational cost. However, it tends to ignore small displacements at the coarse scale. RAFT~\cite{teed2020raft} builds a dense correlation volume, however, limited to memory and computational complexity, its correlation volume is built from 1/8 scale, i.e., a coarse resolution. \cite{jiang2021learning_optical_flow} replace 4D correlation volume with sparse 2D correlation volume to represent the correspondences between frames. Not only use the CNN to estimate the optical flow, after the success of Transformer~\cite{vaswani2017attention} in computer vision, recent methods~\cite{sui2022craft,huang2022flowformer,shi2023flowformer++} utilize Transformer to obtain global dense correlation volume which is still memory-consuming. From the above methods, it seems inevitable that introducing correlation volume will increase computational complexity significantly.
So, in this work, we want to directly predict optical flow from the feature space instead of using the correlation volume.

Due to the camera's and objects' motions, occlusion regions are present in frames, where a fraction of pixels are shown in one frame while invisible in another. Once we warp the features or images based on the estimated optical flow, it will introduce significant visual artifacts in the occluded areas which increase prediction errors.
Following the photometric consistency criterion, several unsupervised methods estimate the occlusion maps to exclude the occluded regions from loss function.
Prevalent methods~\cite{jonschkowski2020matters,stone2021smurf} estimate occlusion maps via forward-backward optical flow consistency. Janai et al.~\cite{janai2018unsupervised} utilize multiple frames for occlusion reasoning. Meister et al.~\cite{meister2018unflow} warp the flow fields to estimate occlusion maps, which raises the question of whether the warping operation will introduce extra prediction errors. UFlow~\cite{jonschkowski2020matters} investigates key components of the unsupervised loss function of optical flow, such as occlusion-aware photometric consistency and so on. Besides, several methods~\cite{meister2018unflow,wang2018occlusion,janai2018unsupervised,luo2021upflow,stone2021smurf} exclude the occluded regions when calculating loss which relies on an accurate occlusion map. 
However, the key factor of solving the inaccuracy caused by the warp operation has not been fully discussed. So, in our method, we combine the occlusion estimation and warping process to reduce errors due to the visual artifacts.
\subsection{Self-supervised Learning}
Self-supervised learning has flourished recently due to its flexibility and good performance. It enables AI systems to learn from orders of magnitude more data, which is important to recognize and understand patterns of subtle, less common representations of the world.
As the self-supervised learning leverages the intrinsic feature, it can get off the struggle of labeling the data.
Meanwhile, the self-supervised learning method will make the network learn the discriminative feature. There are some famous self-supervised learning methods in other tasks such as~\cite{he2020momentum,gidaris2018unsupervised}.
For predicting the optical flow, some researchers tend to develop a semi-supervised or unsupervised framework to estimate optical flow. Both styles benefit from the massive unlabeled real-world datasets. On one hand, it is intuitive to present a semi-supervised~\cite{lai2017semi,im2022semi,yan2020optical} or unsupervised framework~\cite{meister2018unflow,wang2018occlusion,janai2018unsupervised,luo2021upflow,stone2021smurf} trained with more unlabeled data.
Specifically, Lai et al.~\cite{lai2017semi} capture the structural patterns via the adversarial loss without any explicit assumptions. Im et al.~\cite{im2022semi} propose a flow supervisor to fine-tune models with additional unlabeled datasets. Yan et al.~\cite{yan2020optical} use domain transformation to make a pair between foggy and clean images and adaptive estimate the optical flow in fog based on the ground truth of clean images. SMURF~\cite{stone2021smurf} utilizes an unsupervised loss function to train a self-teaching model. However, these methods don't consider the kinetics feature of the optical flow, which still generates the motion information just from the apparent. Prevailing semi-supervised methods utilize unlabeled target domain data with existing labeled synthetic data to train their models. However, there still exists a relatively large performance gap between the supervised and unsupervised methods. And in this work, we will use the kinetics information in the self-supervised process to make the proposed method more robust.
% \vspace{-1em}
\section{Methodology}
\label{sec:methodology}
In this part, we will introduce the proposed method and core insights in the proposed method in detail, and \cref{fig:overview_framework} is the overview of the proposed method.
\vspace{-0.5em}
\subsection{Preliminary}
\vspace{-0.5em}
Before diving into the methodology, we provide some notations used in this paper. Firstly, the pipeline of estimating the optical flow can be formed as:
\begin{equation}  
    f_{<t_0, t_1>} = \Phi(\theta;I_{t_0}, I_{t_1})
\end{equation}
$f_{<t_0, t_1>}$ means the optical flow from $I_{t_0}$ to $I_{t_1}$, and in our proposed method, we focus on the forward optical flow. 
The $\Phi$ is the learnable method to predict the optical flow, the $\theta$ means the parameters of the network to predict the optical flow, the $I_{t_0}$ means the first image, and the $I_{t_1}$ means the second image. 
Secondly, we will use the neural network to extract the feature from RGB images, and generally, we use the $F_{t_0}$ to represent the feature extracted from the $I_{t_0}$. Thirdly, we use the $\hat{F}$ to note the predicted thing, while, the symbol without the hat means the ground truth. And for the occlusion map, we will use $O$ to note. Meanwhile, we use the $\widetilde{I}_{t_1}$ to represent the image generated by the warp operation. So in the following sections, we will introduce the core innovations in our paper.
\subsection{Apparent Information Encoder}
Firstly, the frames are in the color space such as RGB, which is suitable for showing the content to humans instead of exploring other characters, so we design the Apparent Information Encoder(AIE) to extract the high-dimension feature from frames and project the RGB images into the high-level dimension feature space.
For AIE, we hope it can be as general as possible to extract features. In contrast, the limit of optical flow data can not train a general feature encoder from scratch, so we need to use the other pre-trained model to take place the initialization of the AIE. Obviously, the most common method is to use ResNet~\cite{he2016deep} or other famous backbones trained on the ImageNet~\cite{deng2009imagenet}. However, considering the aim of the AIE's feature, it is used for the optical flow estimation. We decide to use the model pre-trained in the geometric image matching (GIM) task instead of image classification.
Geometric image matching aims to find correspondences between frames, which is similar to the aim of optical flow. Specifically, GIM algorithms use geometric information as primary features for matching objects in the non-occluded regions, which is similar to estimating optical flow in the foreground areas with slight motion. Meanwhile, compared to optical flow estimation, the real-world training data of the GIM task is much easier and simpler to obtain, which can make the GIM models more general.  
Introducing GIM into optical flow estimation is beneficial to enhance the generalization of feature representation in the real world. Moreover, the GIM models are trained on much larger real-world datasets, which is beneficial for learning better feature representations.
However, there are still some differences between these two tasks, for example, one difference between GIM and optical flow estimation is that GIM focuses on sparse correspondences, which differs from the dense pixel-wise correspondences in all regions of the optical flow estimation task. So, we still need some optical flow characteristics to supervise the network more suitable for predicting the motion information.
Different from MatchFlow~\cite{dong2023rethinking} which pre-trains feature matching encoder towards feature matching task, our Apparent Information Encoder is initialized based on feature matching weights and then fine-tune towards optical flow datasets, as we suppose that there exists a feature matching tendency between feature matching task and optical flow estimation task.

\begin{figure*}
    \centering
    \includegraphics[width=0.9\textwidth]{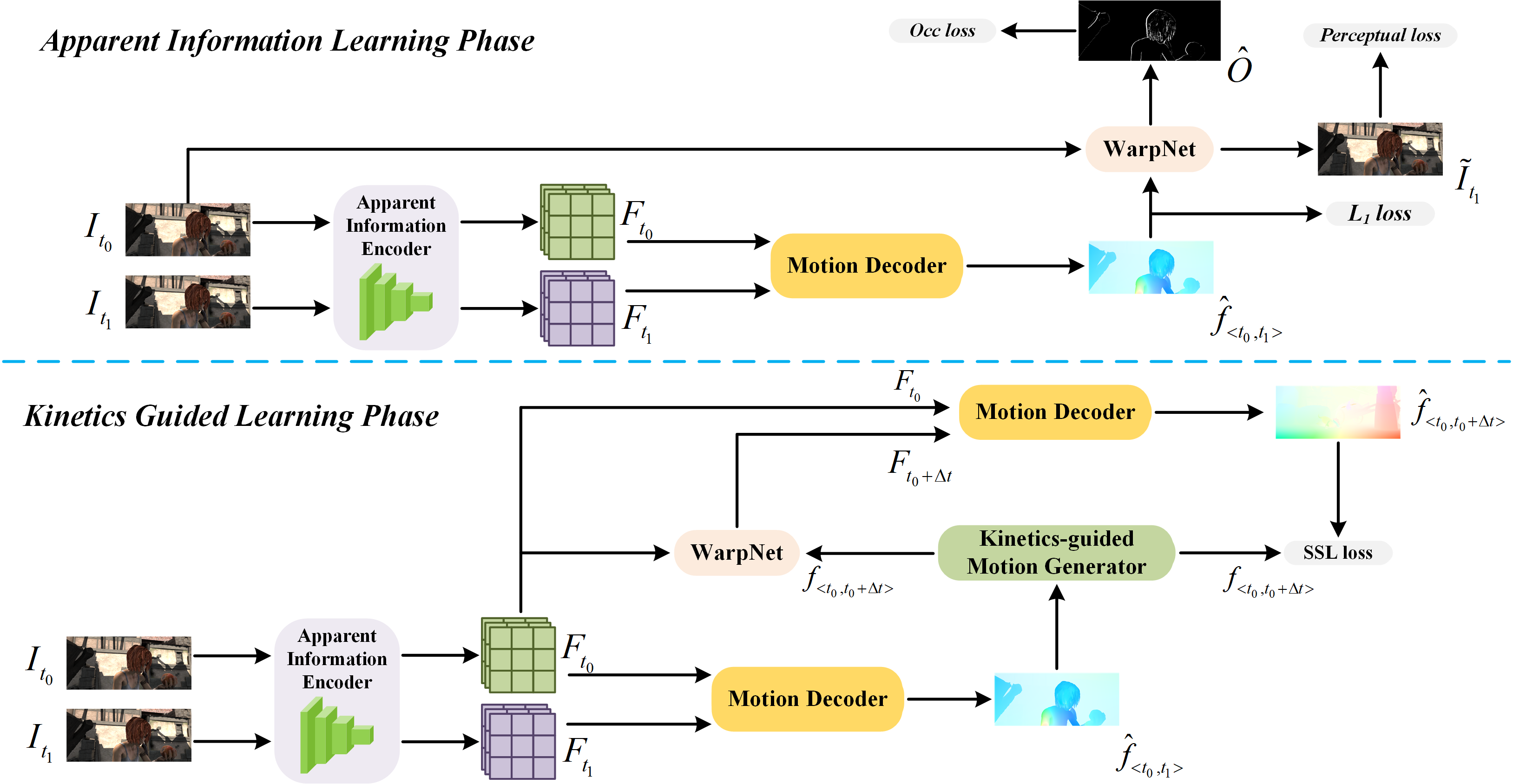}
    \caption{Overview of our proposed method. In the first Apparent Information Learning phase, we will use the labeled data to train the whole network, which is similar to previous methods. In this phase, the network will obtain the ability to get the apparent information. We obtain the predicted occlusion map via WarpNet based on the bi-directional flow output from our proposed Motion Decoder using the perceptual loss and occlusion loss to control training of WarpNet. After that, we obtain the predicted flow $\hat{f}_{<t_0,t_1>}$ and calculate the loss with ground truth. After the Apparent Information Learning phase, we will use the kinetics-guided self-supervised loss to make the network learn the motion information.
    \label{fig:overview_framework}
    }
    \vspace{-1em}
\end{figure*}

\subsection{Feature Space Motion Estimation}
\begin{figure}[htbp]
    \includegraphics[width=1.0\linewidth]{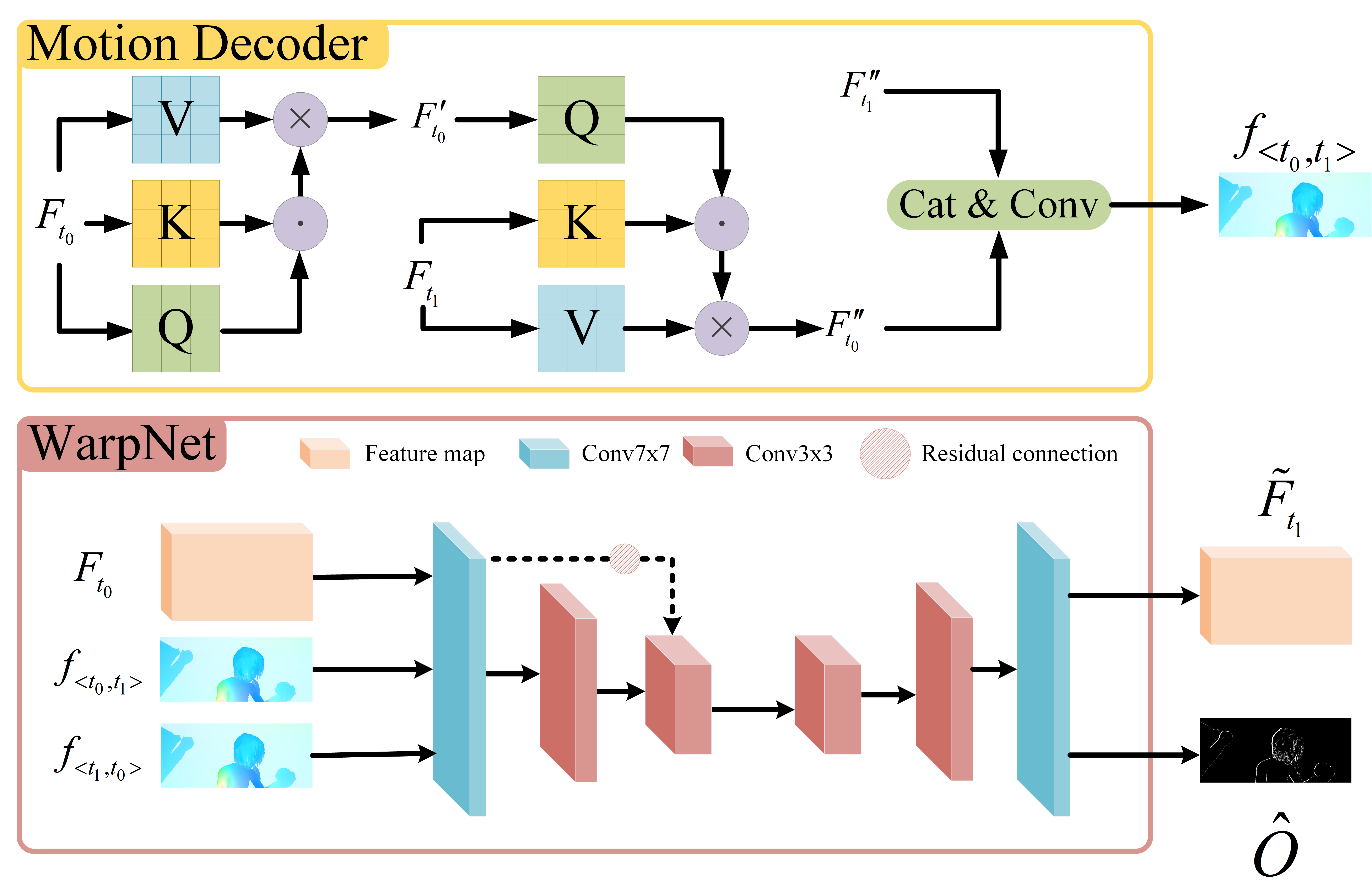}
    \caption{Structure of our Motion Decoder and WarpNet. In the Motion Decoder, we use the $F^{\prime}$ to denote the self-attention result and $F^{\prime\prime}$ to denote the cross-attention result. 
    \label{fig:subnet}
    }
    \vspace{-2em}
\end{figure}

After getting the feature from RGB images, we want to directly use the feature to predict the optical flow instead of using the correlation volume. 
Correlation volume is a prevailing method to construct correspondences of features and it achieves the great performance. However, it increases the usage of computation resources significantly. Motivated by \cite{pan2023drag}, it is intuitive that feature space is sufficiently discriminative to achieve accurate optical flow. It is promising to estimate optical flow from features directly without a correlation volume. Our method directly estimates optical flow from the feature domain instead of building a high computational complexity correlation volume, as shown in \cref{fig:subnet}. 

Considering to achieve the discriminative features and serving for building the relation between features later, we use a Transformer-based~\cite{vaswani2017attention} network to enhance features extracted from apparent information encoder as it benefits from the ability of Transformer to exploit global information and calculate the similarity of two features with the attention mechanism.
For the two consecutive frames, we get the feature, $\mathbf{F}_{t_0}$ and $\mathbf{F}_{t_1}$, respectively. After that, we will add the positional embedding based on channels. The reason why we add the positional information on channels is that we think the importance of each channel is different and the positional information can help the network to learn this kind of importance.
And this kind of global attention also benefits the kinetics-guided self-supervised learning process. After that, we perform stacked self- and cross-attention layers to enhance features. Specifically, the query(Q), key(K), and value(V) of self-attention are the same, while the query is different from the key and value of the cross-attention layer. This process can be formulated by \cref{eq:feature_aug}, where $\mathbf{T}_s$ and $\mathbf{T}_c$ represent a self-attention layer and a cross-attention layer, respectively. 
For the self-attention, we take the KQV as the same input, while for the cross-attention, the Q will be the output of the self-attention, and the K and V are the next frame's feature.
\vspace{-0.5em}
\begin{align}
    \label{eq:feature_aug}
    \mathbf{F}^{\prime\prime}_{t_0} = \mathbf{T}_c(\mathbf{F}^{\prime}_{t_0}, \mathbf{F}^{\mathbf{P}}_{t_1}, \mathbf{F}^{\mathbf{P}}_{t_1}), \mathbf{F}^{\prime}_{t_0} = \mathbf{T}_s(\mathbf{F}^{\mathbf{P}}_{t_0})
\end{align}
As we discard constructing the correlation volume, we find another way to build the relation between extracted features as an auxiliary information $\mathbf{K}_{aux}$ to estimate motion. 
Specifically, we will compute the channel-wise similarity between these two consecutive frames' feature and then concatenate the Top-K similarity channel as the sparse correlation of these two frames, which is denoted by $\mathbf{K}_{aux}$.
To estimate the motion between features, we concatenate the $\mathbf{F}^{\prime\prime}_{t_1}$, $\mathbf{F}^{\prime\prime}_{t_0}$, and $\mathbf{K}_{aux}$ to further enhance the features. We stack residual blocks to estimate optical flow from concatenated features, and for simplicity, we only draw the \textit{concatenate} operation in the~\cref{fig:subnet}. 

\subsection{Learnable Warp Operation}
The motion decoder can decode the optical flow from the feature space, and then we need a warp operation to generate the following time period image, which is vital in the training process. 
So this part will introduce the proposed learnable warp operation, WarpNet. Specifically, we implement a learning-based warping network using a predicted occlusion map as auxiliary information to map motion trajectories, as shown in \cref{fig:subnet}, which introduces a bi-directional estimation style to acquire the occlusion map. To our knowledge, estimating occlusion maps and warping are two separate parts in existing learning-based methods, which means they are not unified optimized yet. 

Significant visual artifacts exist when warping on the occluded areas~\cite{wang2018occlusion,lu2020devon} which increases the error for optical flow estimation. To obtain better warped images or features, we need a precise occlusion map.  
Occlusion is a key challenge in existing optical flow estimation methods. Some methods use a truncated penalty to reduce the effects of the outlier pixels that violate brightness consistency. However, it might raise false-positive matches and propagate the error in the following estimation. So it is necessary to achieve a precise occlusion map for optical flow estimation. 

In our method, we utilize a general approach, leveraging forward-backward motion consistency, to predict the occluded regions. In addition, warping can be viewed as a process of the image or feature transformation, mapping $\mathbf{F}_{t_0}$ to $\mathbf{F}_{t_1}$ based on the provided motion trajectory. Based on a given flow, most frameworks implement warping via $grid\_sample$\cite{jaderberg2015spatial}. However, it lacks an occlusion map to determine which pixel should be filled in the occluded areas. So, we incorporate the occlusion estimation into the warping operation to reduce the error caused by the visual artifacts.

\subsection{Kinetics Self-supervised Learning}
\label{subsec:self-supervised}
This part will introduce the proposed kinetics-guided self-supervised training strategy.
As we have discussed, the previous methods provide great insights from apparent to estimate the optical flow. Nevertheless, as the apparent-based methods require the ground truth to guide the network learning the weights of projecting the image to optical flow, or in other words, the motion information, we cannot easily use the non-labeled data. In this paper, we propose another way to leverage these non-labeled data from the kinetics opinion, which will avoid labeling the optical flow between frames. 

Before describing the proposed methods in detail, we want to review the basic knowledge of kinetics. The uniform linear motion is the basic motion in the kinetics, and if we take the $\Delta{t}$ as the infinitesimal time period, we can think every motion is a uniform linear motion. Therefore, the kinetics characteristics are the same in two consecutive infinitesimal time periods.
So, how about using the characteristics of the motion to guide the network to estimate? As for the other self-supervised algorithm using rotation\cite{gidaris2018unsupervised}, mask~\cite{feichtenhofer2022masked} and so on, we want to use the speed in the motion to construct the self-supervised task. As we have mentioned, in every infinitesimal time period $\Delta{t}$, the speed is a constant in both magnitude and direction. And considering one direction, for example, the horizontal direction $x$, and in each infinitesimal time period, we have:
\begin{equation}
    \mathbf{V}_{x}=\lim _{\Delta t \rightarrow 0} \frac{\Delta \mathbf{x}}{\Delta t}
\end{equation}
Accordingly, in two consecutive infinitesimal time periods, the displacements are the same $\mathbf{V}_{x} \Delta t$. For the optical flow describing the displacement, we can easily get the \cref{eq:kinetic_flow}. And $\alpha$ could be any arbitrary value which is smaller than $1$, for simplicity, we choose the $\alpha$ equals to $0.5$.
\begin{equation}
    f_{<t_0, t_0+\alpha\Delta t>} = \alpha f_{<t_0,t_0+\Delta t>}
    \label{eq:kinetic_flow}
\end{equation}
% In stage II, we finetune our model on the unlabeled datasets DAVIS\cite{Caelles_davis_2019} and High Speed Sintel\cite{janai2017slow} for 100K iterations.

So, we can use this basic characteristic in the motion as the self-supervised information to guide the network. Meanwhile, the time period between two consecutive frames usually is minimal, for example, as the FPS is 30, the time period between frames is around 33ms.
As shown in \cref{fig:overview_framework}, we combine the apparent and kinetics through the self-supervised loss. We computed the kinetics-guided self-supervised loss between the teacher and student flow. The teacher flow, $f_{<t_0,t_0+\Delta t>}$, is generated by the Kinetics-guided Motion Generator based on the constant velocity priors through~\cref{eq:kinetic_flow}. And the student flow, $\hat{f}_{<t_0,t_0+\Delta t>}$, is produced by the Motion Decoder through the $\mathbf{F}_{t_0}$ and $\mathbf{F}_{t_0+\Delta t}$ generated by the WarpNet.The loss function is shown in~\cref{eq:unsupervised_loss}. 
% \vspace{-2em}
\begin{equation}
    \label{eq:unsupervised_loss}
    % \begin{split}
        % f_{t\rightarrow {t+\Delta t}} &= f_{t\rightarrow t+1} \times \Delta t \\
    \mathbf{L}_{kinetics} = \sum_{i=1}^{N}\gamma^{N-i}||f_{<t_0,t_0+\Delta t>}-\hat{f}_{<t_0,t_0+\Delta t>}||_1 
    % \end{split}
\end{equation}

\subsection{Loss Function}
This part will introduce the loss function used in the proposed methods. 
As we have mentioned, the whole network's training will be split into two phases, and each phase will have some different loss function to teach the network's training process. Firstly, in the Apparent Information Learning(AIL) phase as shown in~\cref{fig:overview_framework}, we train our model using the supervised learning strategy. Similar to RAFT\cite{teed2020raft}, we use $\mathbf{L}_1$ loss between predicted flow and ground truth to supervise training, with exponentially increasing weights $\gamma$, defined as \cref{eq:sl_loss}.
\vspace{-0.5em}
\begin{equation}
    \mathbf{L}_{1} = \sum_{i=1}^{N}\gamma^{N-i}||f-\hat{f}_i||_1
    \label{eq:sl_loss}
    \vspace{-0.5em}
\end{equation}
For training WarpNet, we introduce two losses. One is a perceptual loss between the original image $I$ and the warped image $\widetilde{I}$ computed by the pre-trained VGG-19\cite{johnson2016perceptual} network, defined as \cref{eq:perceptual_loss} where $V_i$ is the $i^{th}$ layer of the pre-trained VGG-19 network, and $j$ represents that the image is downsampled $j$ times\cite{zhao2022thin}.
% \vspace{-1em}
\begin{equation}
    \mathbf{L}_{perceptual} = \sum_{j}\sum_{i}||V_i(I_j) - V_i(\widetilde{I}_j)||_1
    \label{eq:perceptual_loss}
    \vspace{-0.5em}
\end{equation}
Another loss is the $L_1$ loss of the occlusion map:
\begin{equation}
    \mathbf{L}_{occ} = ||O-\hat{O}||_1
    \label{eq:occ_loss}
    \vspace{-0.5em}
\end{equation}
Overall, the loss function used in AIL is \cref{eq:ail_loss}.
\begin{equation}
    \mathbf{L}_{AIL} = \mathbf{L}_{1} + \mathbf{L}_{perceptual} + \mathbf{L}_{occ}
    \label{eq:ail_loss}
\end{equation}
For the Kinetics Guided Learning(KGL) phase, we use the self-supervised loss introduced in \cref{subsec:self-supervised}. Meanwhile, we also use the perceptual loss on the WarpNet to increase the performance. So, the loss function is defined in \cref{eq:kgl_loss}. The more detailed description of these mentioned loss functions is provided in the supplementary.
\begin{equation}
    \mathbf{L}_{KGL} = \mathbf{L}_{kinetics} + \mathbf{L}_{perceptual}
    \label{eq:kgl_loss}
\end{equation}
% \vspace{-2em}
\section{Experiments}
\label{sec:exps}

%% tab: eval on train+test set
\begin{table*}[!b]
    \centering
    \vspace{-1em}
    \caption{Results on the Sintel and KITTI-2015. \textsuperscript{*} represents the results of warm-start strategy used in RAFT\cite{teed2020raft}. $\dagger$ denotes the unsupervised methods that are trained using images from the target domain. $\ddagger$ represents the semi-supervised learning framework. The Gain/Param. measures the parameter gains of Fl-all on KITTI-2015 test set and $\downarrow$ indicates lower is more efficient. The best results are \textbf{bolded} and the second best results are \underline{underlined}.\label{table:eval_testset}}
    \vspace{-1em}
    \resizebox{0.8\linewidth}{!}{
    \begin{tabular}{l |c cc c cc c c c cc}
    % \hline
    \toprule
    \multirow{2}{*}{Method} && \multicolumn{2}{c}{Sintel(train)} && \multicolumn{2}{c}{KITTI-15(train)} && KITTI-15(test) && Params. & \multirow{2}{*}{Gain/Param.($\downarrow$)} \\ \cline{3-4} \cline{6-7} \cline{9-9} 
    \multicolumn{1}{c|}{}   && Clean(EPE) & Final(EPE) && EPE & Fl-all && Fl-all && (M) & \multicolumn{1}{c}{}\\ 
    \midrule
    UpFlow\cite{luo2021upflow}\textsuperscript{$\dagger$} && 2.33 & 2.67 && 2.45 & - && 9.38 && 3.5 & 2.69 \\
    Lai et al.\cite{lai2017semi}\textsuperscript{$\ddagger$} && 2.41 & 3.16 && 14.69 & 30.30 && 31.01 && - & - \\
    RAFT\cite{teed2020raft} && 0.76/0.77\textsuperscript{*} & 1.22/1.27\textsuperscript{*} && 0.63 & 1.50 && 5.10 && 5.3 & 0.97 \\
    SCV\cite{jiang2021learning_optical_flow}&& 0.79/0.86\textsuperscript{*} & 1.70/1.75\textsuperscript{*} && 0.75 & 2.10 && 6.17 && 5.3 & 1.16 \\
    GMFlow\cite{xu2022gmflow}               && - & - && - & - && 9.32 && 4.7 & 1.99\\
    MatchFlow(G)\cite{dong2023rethinking}   && \textbf{0.49} & \textbf{0.78} && \textbf{0.55} & \textbf{1.10} && \textbf{4.63} && 15.4 & 0.30\\
    Ours                                    && \underline{0.50} & \underline{0.87} && \textbf{0.55} & \underline{1.23} && \underline{4.94}   && 11.8 & 0.42 \\
    \bottomrule
    \end{tabular}
    }    
\end{table*}
%kitti vis
\begin{figure*}[t]
    \centering
    \vspace{-1em}
    \includegraphics[width=0.8\linewidth]{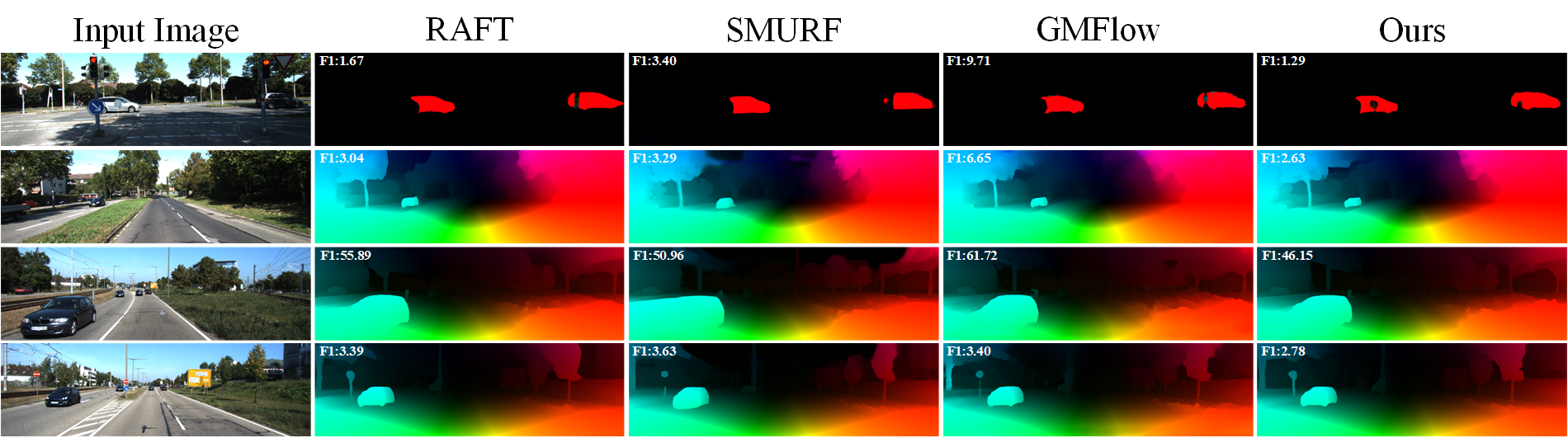}
    \caption{Qualitative comparison of ours with the state-of-the-art methods on KITTI-2015 test set.\label{fig:fig3_vis_kitti}}
    \vspace{-2em}    
\end{figure*}

\subsection{Setting up}
\textbf{Datasets.} We firstly choose some widely used datasets, such as FlyingChairs~\cite{Dosovitskiy_FlowNet}, FlyingThings3D~\cite{Mayer_A_Large_Dataset}, MPI Sintel~\cite{Butler_A_Naturalistic}, and KITTI-2015~\cite{Menze2015CVPR}. 
FlyingChairs~\cite{Dosovitskiy_FlowNet} provides \num{22962} samples which are split into \num{22322} and \num{640} for training and validation. FlyingThings3D~\cite{Mayer_A_Large_Dataset} generates \num{107040} ground truth including \num{89560} for training and \num{17480} for validation from different views and directions, i.e., to the future or the past. Sintel~\cite{Butler_A_Naturalistic} is a synthetic dataset derived from the 3D animated film Sintel, which provides \num{1041} samples for training and \num{552} samples for testing. Both FlyingThings3D and Sintel provide \textit{cleanpass} and \textit{finalpass}. And in the \textit{finalpass}, it is more challenging to estimate precise optical flow as there exist some effects, e.g., motion blur, atmospheric effects, specular reflections, and so on. KITTI-2015~\cite{Menze2015CVPR} presents a real-world dataset for autonomous driving, which provides \num{200} samples of sparse optical flow with corresponding object maps. Except for KITTI, the other three datasets are synthetic datasets, which are different from real-world scenes.
We also introduce extra datasets to leverage their kinetics information, i.e., HD1K~\cite{kondermann2016hci} is a high resolution dataset usually used for fine-tuning Sintel; High Speed Sintel~\cite{janai2017slow} is a high framerate dataset with low motion magnitude, around 99\% pixels falling into 0-10 pixels; DAVIS~\cite{Caelles_davis_2019} is a real-world dataset and we use it to reduce the gap between synthetic and real-world scenes.

\noindent \textbf{Metrics.} We use the End-Point-Error (EPE) and Fl-all(\%) metrics for evaluation. EPE calculates the average $\mathbf{L}_2$ distance between the predicted and ground truth optical flow. To better understand the performance gains, we also report the EPE in different motion magnitudes, denoted as $s_{0-10}$, $s_{10-40}$, and $s_{40+}$, measuring the EPE over all of ground truth pixels with motion magnitude falling to $0-10$, $10-40$, and more than 40 pixels, respectively. Fl-all~\cite{Menze2015CVPR} measures the percentage of outliers averaged over all ground truth pixels, defined in \cref{eq:fl} whose $N$ is the number of valid pixels.
\begin{equation}
%Fl-all
    \label{eq:fl}
    \begin{split}
        Fl_{all} &= \frac{|\{ p_{i,j} | EPE_{p_{i,j}} > 3, \frac{EPE_{p_{i,j}}}{mag^{gt}_{p_{i,j}}} >0.05\}|}{N} \\
        mag^{gt}_{p_{i,j}} &= \sqrt{u^2_{p_{i,j}}+v^2_{p_{i,j}}}, i\in[0,H), j\in[0,W) \\
    \end{split}
\end{equation}

%% training and evaluation settings
\noindent \textbf{Implementation.} 
Our method is implemented in PyTorch. We use the AdamW\cite{loshchilov2018decoupled} optimizer with a one-cycle learning rate policy. 
In the Apparent Information Learning phase, following the ~\cite{Dosovitskiy_FlowNet,ilg2017flownet}, we first train the model on FlyingChairs dataset for 150K iterations, with a batch size of 16 and a learning rate of 4e-4, then train for 400K iterations on FlyingThings3D with a batch size of 8 and a learning rate of 2e-4. After that, we follow the RAFT~\cite{teed2020raft} to fine-tune our model on Sintel\cite{Butler_A_Naturalistic} and KITTI-2015\cite{Menze2015CVPR}. 
In the Kinetics Guided Learning phase, we use the self-supervised kinetics loss on unlabeled High Speed Sintel\cite{janai2017slow} and DAVIS\cite{Caelles_davis_2019} for another 100K iterations using the model fine-tuned on KITTI. The more details about the training and inference process are in the supplement. 
\subsection{Results}
\textbf{Quantitative analysis.} \cref{tab:comparison_gmflow_raft_ours} provides detailed comparisons of ours with the state-of-the-art methods in different motion magnitude areas. From this table, we know that the proposed method can outperform most SOTA methods.
We evaluate the results on the training set of Sintel\cite{Butler_A_Naturalistic} and KITTI-2015\cite{Menze2015CVPR} to make a quantitative comparison of our method and the state-of-the-arts.
% These models are pre-trained on FlyingChairs and FlyingThings3D. 

%% tab: eval on training set
\begin{table}[htbp]    
    \centering
    \vspace{-0.5em}
    \caption{Quantitative results on training set of Sintel and KITTI-2015. Models are pre-trained on FlyingChairs(C) and FlyingThings3D(T). The best results are \textbf{bolded}, and the second best results are \underline{underlined}.\label{table:result_cross_evaluation}}
    \vspace{-1em} 
    \resizebox{1.0\linewidth}{!}{
    \begin{tabular}{ l| c cc c cc }
    \toprule
    \multirow{2}{*}{Method} && \multicolumn{2}{c}{Sintel(train)} && \multicolumn{2}{c}{KITTI-15(train)} \\ \cline{3-4} \cline{6-7}
    \multicolumn{1}{c|}{}   && Clean(EPE) & Final(EPE) && EPE  & Fl-all \\ 
    \midrule
    RAFT(small)\cite{teed2020raft}         && 2.21  & 3.35  && 7.51 & 26.90 \\
    RAFT(2-view)\cite{teed2020raft}        && 1.43  & 2.71  && 5.04 & 17.46 \\
    GMA\cite{jiang2021learningtoestimate}  && 1.30  & 2.74  && \underline{4.69} & 17.10 \\
    SCV\cite{jiang2021learning_optical_flow}&&1.29  & 2.95  && 6.80 & 19.30  \\ 
    GMFlow\cite{xu2022gmflow}              && \underline{1.08}  & \underline{2.48}  && 7.77 & 23.40 \\
    % MatchFlow(R)\cite{dong2023rethinking}  && 1.14  & 2.61  && 4.19 & \textbf{13.60} \\
    MatchFlow(G)\cite{dong2023rethinking}  && \textbf{1.03}  & \textbf{2.45}  && \textbf{4.08} & \underline{15.60} \\
    % AnyFlow\cite{jung2023anyflow}          && 1.17  & 2.58  && \underline{3.95} & \textbf{13.01} \\
    Ours                             && 1.17  & 2.50     &&  6.34   &  \textbf{15.52}   \\ 
    \bottomrule
    \end{tabular}
    }
    \vspace{-1em} 
\end{table}

%more metrics
\begin{table*}[!b]
    \centering
    \caption{Comparison of ours with SOTA methods on more metrics. The best results are \textbf{bolded} and the second best results are \underline{underlined}.\label{tab:comparison_gmflow_raft_ours}} 
    \vspace{-1em} 
    \resizebox{1.0\linewidth}{!}{
    \begin{tabular}{l|c cccc c cccc c ccccc}
        \toprule
        \multirow{2}{*}{Method}  && \multicolumn{4}{c}{Sintel(train, clean)} && \multicolumn{4}{c}{Sintel(train, final)} && \multicolumn{5}{c}{KITTI-15(train)}\\
        \cline{3-6} \cline{8-11} \cline{13-17}
        \multicolumn{1}{l|}{}       && EPE & $s_{0-10}$ & $s_{10-40}$ & $s_{40+}$ && EPE & $s_{0-10}$ & $s_{10-40}$ & $s_{40+}$ && EPE & $s_{0-10}$ & $s_{10-40}$ & $s_{40+}$ & Fl-all \\ 
        \midrule
        RAFT\cite{teed2020raft}&& 0.769 & 0.193 & 0.870 & 4.629 && 1.218 & 0.266 & 1.317 & 7.753 && 0.640 & 0.147 & 0.389 & 1.364 & 1.517 \\
        GMA\cite{jiang2021learningtoestimate}&& \underline{0.617} & \underline{0.171} & \underline{0.755} & 3.475 && \underline{1.061} & \underline{0.226} & \underline{1.160} & 6.765 && \underline{0.566} & \underline{0.118} & \underline{0.336} & \underline{1.228} & \textbf{1.223} \\
        GMFlow\cite{xu2022gmflow}   && 0.762 & 0.380 & 0.856 & \underline{3.266} && 1.110 & 0.433 & 1.203 & \underline{5.702} && 1.361 & 0.202 & 0.425 & 3.451 & 5.167 \\
        Ours                        && \textbf{0.499} & \textbf{0.125} & \textbf{0.578} & \textbf{2.981}  && \textbf{0.872} & \textbf{0.183} & \textbf{0.935} & \textbf{5.619} && \textbf{0.552} & \textbf{0.103} & \textbf{0.314} & \textbf{1.225} & \underline{1.232}\\
        \bottomrule
    \end{tabular}
    }
\end{table*}
As shown in \cref{table:result_cross_evaluation}, for Sintel cleanpass, our method improves RAFT(2-view) by 18\% (from 1.43 to 1.17). On the Sintel finalpass, our result is comparable to all the supervised state-of-the-art methods, currently behind GMFlow~\cite{xu2022gmflow} and MatchFlow~\cite{dong2023rethinking} while better than all the other methods. On the KITTI-2015 dataset, our result of EPE is better than GMFlow~\cite{xu2022gmflow} and only behind MatchFlow~\cite{dong2023rethinking} on the Fl-all metric. 
\cref{table:eval_testset} reports the results on the Sintel\cite{Butler_A_Naturalistic} training set and KITTI-2015\cite{Menze2015CVPR}. Obviously, our method outperforms the most existing unsupervised and semi-supervised methods. Specifically, For the training set, on the Sintel cleanpass, we improve the EPE by 78\% and 79\% compared to \cite{luo2021upflow} and \cite{lai2017semi} (from 2.33 and 2.41 to 0.50). On the Sintel finalpass, our method achieves EPE 0.87 which is only behind MatchFlow~\cite{dong2023rethinking}. For KITTI-2015, EPE and Fl-all of our method are 0.55 and 1.23\% which is comparable to state-of-the-art frameworks. For the test set of KITTI-2015, our method reduces the Fl-all value of 9.38\% in UpFlow\cite{luo2021upflow} to 4.94\% with around 47\% improvements.

\noindent \textbf{Visualization analysis.} We also provide our visual comparison and the state-of-the-art methods in \cref{fig:fig3_vis_kitti}. The results of visualization prove that our method can achieve better visual performance on object edges.

% \vspace{-1em}
\vspace{-0.5em}
\section{Ablation Study}
\label{sec:ablations}
\vspace{-0.5em}
\noindent In this part, we conducted the ablation experiments to validate the efficiency of our proposed method. And the more ablation studies are in the supplement.
%innovations
\begin{table*}[t]
    \centering
    \caption{Ablation studies on the contribution of innovations on Sintel and KITTI-2015 training set.\label{tab:ablation_studies_inno}}
    \vspace{-1em}
    \resizebox{1.0\linewidth}{!}{
        \centering
    \begin{tabular}{l|c cccc c cccc c ccccc}
        \toprule
        \multirow{2}{*}{Method}  && \multicolumn{4}{c}{Sintel(train, clean)} && \multicolumn{4}{c}{Sintel(train, final)} && \multicolumn{5}{c}{KITTI-15(train)} \\
        \cline{3-6} \cline{8-11} \cline{13-17}
        \multicolumn{1}{l|}{}       && EPE & $s_{0-10}$ & $s_{10-40}$ & $s_{40+}$ && EPE & $s_{0-10}$ & $s_{10-40}$ & $s_{40+}$ && EPE & $s_{0-10}$ & $s_{10-40}$ & $s_{40+}$ & F1-all \\ 
        \midrule
        Full Model                  && 0.499 & 0.125 & 0.578 & 2.981 && 0.872 & 0.183 & 0.935 & 5.619 && 0.552 & 0.103 & 0.314 & 1.225 & 1.232 \\
        w/o Pre-trained GIM         && 0.680 & 0.147 & 0.713 & 4.384 && 1.273 & 0.241 & 1.274 & 8.587 && 0.616 & 0.116 & 0.337 & 1.378 & 1.524 \\
        w/o WarpNet                 && 0.593 & 0.138 & 0.654 & 3.685 && 1.081 & 0.201 & 1.095 & 7.288 && 0.580 & 0.106 & 0.321 & 1.298 & 1.372 \\
        w/o Self-supervised Loss&& 0.786 & 0.157 & 0.843 & 5.124 && 1.446 & 0.227 & 1.394 & 10.206&& 0.789 & 0.151 & 0.417 & 1.781 & 2.292 \\
        \bottomrule
    \end{tabular}
    }
\end{table*}

% Without specific notation, the following ablation studies are performed with models pre-trained on FlyingChairs and FlyingThings3D.

\noindent\textbf{Innovations analysis.} \cref{tab:ablation_studies_inno} shows the contributions of each innovation on Sintel and KITTI-2015, respectively. Obviously, with the self-supervised loss to fine-tune our model, the performance increases significantly, which proves the effectiveness of the proposed kinetics self-supervised manner. Meanwhile, the other proposed modules also improve the performance. Specifically, the improvement of using the pre-trained GIM model demonstrates that the general feature extractor is useful in generating discriminative features to help low-level computer vision tasks, such as optical flow estimation. And the great performance of the WarpNet shows the potential that we can use a differentiable module to realize the traditional warp operation. 
%ss gains
\begin{figure}[h]
    \includegraphics[width=0.9\linewidth]{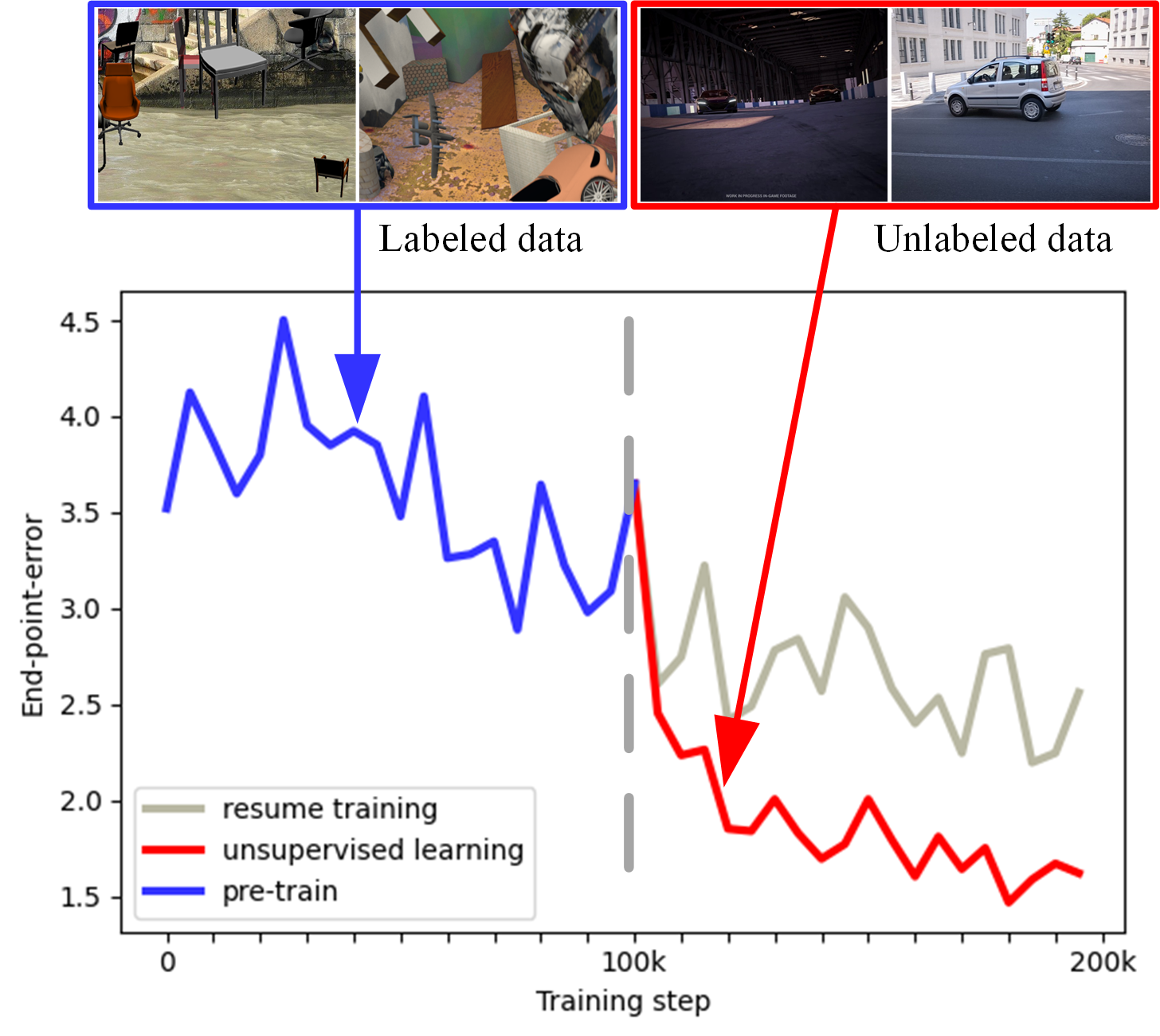}
    \caption{EPE on Sintel training set. Our model uses more unlabeled datasets DAVIS\cite{Caelles_davis_2019} and High Speed Sintel\cite{janai2017slow} to fine-tune our model. Our method decreases more loss compared to the supervised style with the same iterations.\label{fig:fig4_ss_gains}}
    \vspace{-2.5ex}
\end{figure}

\noindent\textbf{Self-supervised Learning.} \cref{fig:fig4_ss_gains} shows using a self-supervised strategy to fine-tune our model achieves lower EPE on Sintel training set. The first 100K steps are trained on FlyingThings3D. The training datasets on the latter 100K steps of unsupervised learning and the resume training are DAVIS\cite{Caelles_davis_2019}, High Speed Sintel\cite{janai2017slow} and FlyingThings3D.

\noindent\textbf{Pre-trained GIM.} We make an experiment to select which pre-trained feature matching model is better for our method, as shown in \cref{tab:ablation_studies_pretrained_models}. We train these models for 150K iterations on the FlyingChairs dataset and evaluate on the validation set of FlyingChairs. Finally, we select the LoFTR(outdoor) as our pre-trained Apparent Information Encoder.
% LoFTR~\cite{sun2021loftr} and Quadtree Attention-based LoFTR~\cite{tang2021quadtree} are state-of-the-art methods in feature matching. We select their best pre-trained models as initialization of our feature extractor and make an ablation study to decide the final model in \cref{sec:ablations}. 
% pretrained GIM models
\begin{table}[htbp]
    \centering
    \caption{Ablations on pre-trained feature matching models. The best results are \textbf{bolded}. $\diamond$ represents the default settings. 1px, 3px, and 5px represent the fractions of EPE over those, respectively.\label{tab:ablation_studies_pretrained_models}}
    \resizebox{.9\linewidth}{!}{
    \begin{tabular}{l|c cccc}
        \toprule
        \multirow{2}{*}{Model} && \multicolumn{4}{c}{FlyingChairs}  \\ \cline{3-6}
        \multicolumn{1}{l|}{} && EPE & 1px & 3px & 5px \\
        \midrule
        LoFTR(indoor)~\cite{sun2021loftr}         && 0.997 & 0.131 & 0.047 & 0.031 \\
        LoFTR(outdoor)~\cite{sun2021loftr}\textsuperscript{$\diamond$}&& \textbf{0.866} & \textbf{0.127} & \textbf{0.042} & \textbf{0.027} \\
        QTA(indoor)~\cite{tang2021quadtree}           && 11.321 & 0.772 & 0.619 & 0.515 \\
        QTA(outdoor)~\cite{tang2021quadtree}          && 11.321 & 0.772 & 0.619 & 0.515 \\
        Scratch               && 1.096  & 0.157 & 0.056 & 0.036 \\
        \bottomrule
    \end{tabular}
    }
    \vspace{-1em}
\end{table}

\noindent\textbf{GPU memory usage.} \cref{tab:ablation_studies_gpu_memory} shows that although the number of parameters of our model is around double of RAFT~\cite{teed2020raft}, our method consumes the least computational time, only 55ms for a single pair of images in the FlyingChairs dataset. The GRU iteration number of RAFT~\cite{teed2020raft}, GMA\cite{jiang2021learningtoestimate}, and MatchFlow~\cite{dong2023rethinking} is set to 32 equally for comparison.

%%% params, GPU memory
\begin{table}[htbp]
    \centering
    \caption{Comparison of parameters, GPU memory, and time.\label{tab:ablation_studies_gpu_memory}}
    \vspace{-1em}
    \resizebox{1.0\linewidth}{!}{
    \begin{tabular}{l|c c c c cc}
    \toprule
    \multirow{2}{*}{Method} & Params & Memory & Time & \multirow{2}{*}{$\Delta\frac{Memory(GB)}{Params(M)}$} & KITTI-15(test) \\ \cline{6-6}
    \multicolumn{1}{c|}{} & (M) & (GB) & (ms/pair) & \multicolumn{1}{c}{} & Fl-all \\
    \midrule
    RAFT\cite{teed2020raft}  & 5.26 & \textbf{10.57} & 155 & 2.01  & 5.10\\
    GMA\cite{jiang2021learningtoestimate} & 5.88 & 13.46 & 99 & 2.29  & 5.15\\
    GMFlow\cite{xu2022gmflow}  & \textbf{4.68} & 13.27 & 167 & 2.84  & 9.32 \\
    Ours & 11.81 & 16.93 & \textbf{55} & \textbf{1.43} & \textbf{4.94}\\
    \bottomrule
    \end{tabular}
    }
\end{table}

% \vspace{-1em}
\section{Conclusion}
\label{sec:conclusion}
This work proposes a kinetics-guided self-supervised loss function to train the optical flow estimation model. In the proposed method, we directly predict the motion information from the high-dimension semantic feature instead of using the dense correlation volume. Meanwhile, this paper proposes a new warp operation that can be optimized with the whole network. Specifically, discarding the correlation volume improves the efficiency of the proposed, and the WarpNet alleviates the error caused by the occlusion. Extensive experimental results show that our method achieves state-of-the-art performance. Furthermore, we think the most important part of this paper is about using the kinetics in a self-supervised manner, starting from the motion's basic properties. By using this self-supervised manner, researchers can use countless data to improve their optical flow models in their situations, and experiment results prove promising to research the self-supervised optical flow estimation methods. So, in the future, we will continue exploring this manner and propose a new method without using labeled data to train the optical flow method, achieving state-of-the-art performance.    

\clearpage
{
    \small
    \bibliographystyle{unsrt}
    \bibliography{main}

\begin{thebibliography}{10}

\bibitem{brox2010large}
Thomas Brox and Jitendra Malik.
\newblock Large displacement optical flow: descriptor matching in variational
  motion estimation.
\newblock {\em PAMI}, 33(3):500--513, 2010.

\bibitem{weinzaepfel2013deepflow}
Philippe Weinzaepfel, Jerome Revaud, Zaid Harchaoui, and Cordelia Schmid.
\newblock Deepflow: Large displacement optical flow with deep matching.
\newblock In {\em ICCV}, 2013.

\bibitem{revaud2015epicflow}
Jerome Revaud, Philippe Weinzaepfel, Zaid Harchaoui, and Cordelia Schmid.
\newblock Epicflow: Edge-preserving interpolation of correspondences for
  optical flow.
\newblock In {\em CVPR}, 2015.

\bibitem{bailer2015flow}
Christian Bailer, Bertram Taetz, and Didier Stricker.
\newblock Flow fields: Dense correspondence fields for highly accurate large
  displacement optical flow estimation.
\newblock In {\em ICCV}, 2015.

\bibitem{hur2017mirrorflow}
Junhwa Hur and Stefan Roth.
\newblock Mirrorflow: Exploiting symmetries in joint optical flow and occlusion
  estimation.
\newblock In {\em ICCV}, 2017.

\bibitem{hosni2012fast}
Asmaa Hosni, Christoph Rhemann, Michael Bleyer, Carsten Rother, and Margrit
  Gelautz.
\newblock Fast cost-volume filtering for visual correspondence and beyond.
\newblock {\em PAMI}, 35(2):504--511, 2012.

\bibitem{Dosovitskiy_FlowNet}
Alexey Dosovitskiy, Philipp Fischer, Eddy Ilg, Philip Häusser, Caner Hazirbas,
  Vladimir Golkov, Patrick van~der Smagt, Daniel Cremers, and Thomas Brox.
\newblock Flownet: Learning optical flow with convolutional networks.
\newblock In {\em ICCV}, 2015.

\bibitem{ilg2017flownet}
Eddy Ilg, Nikolaus Mayer, Tonmoy Saikia, Margret Keuper, Alexey Dosovitskiy,
  and Thomas Brox.
\newblock Flownet 2.0: Evolution of optical flow estimation with deep networks.
\newblock In {\em CVPR}, 2017.

\bibitem{sun2018pwc}
Deqing Sun, Xiaodong Yang, Ming-Yu Liu, and Jan Kautz.
\newblock Pwc-net: Cnns for optical flow using pyramid, warping, and cost
  volume.
\newblock In {\em CVPR}, 2018.

\bibitem{teed2020raft}
Zachary Teed and Jia Deng.
\newblock Raft: Recurrent all-pairs field transforms for optical flow.
\newblock In {\em ECCV}, 2020.

\bibitem{jiang2021learning_optical_flow}
Shihao Jiang, Yao Lu, Hongdong Li, and Richard Hartley.
\newblock Learning optical flow from a few matches.
\newblock In {\em CVPR}, 2021.

\bibitem{brox2004highacc}
Thomas Brox, Andr{\'e}s Bruhn, Nils Papenberg, and Joachim Weickert.
\newblock High accuracy optical flow estimation based on a theory for warping.
\newblock In {\em ECCV}, 2004.

\bibitem{ranjan2017optical}
Anurag Ranjan and Michael~J Black.
\newblock Optical flow estimation using a spatial pyramid network.
\newblock In {\em CVPR}, 2017.

\bibitem{vaswani2017attention}
Ashish Vaswani, Noam Shazeer, Niki Parmar, Jakob Uszkoreit, Llion Jones,
  Aidan~N Gomez, {\L}ukasz Kaiser, and Illia Polosukhin.
\newblock Attention is all you need.
\newblock In {\em NIPS}, 2017.

\bibitem{sui2022craft}
Xiuchao Sui, Shaohua Li, Xue Geng, Yan Wu, Xinxing Xu, Yong Liu, Rick Goh, and
  Hongyuan Zhu.
\newblock Craft: Cross-attentional flow transformer for robust optical flow.
\newblock In {\em CVPR}, 2022.

\bibitem{huang2022flowformer}
Zhaoyang Huang, Xiaoyu Shi, Chao Zhang, Qiang Wang, Ka~Chun Cheung, Hongwei
  Qin, Jifeng Dai, and Hongsheng Li.
\newblock Flowformer: A transformer architecture for optical flow.
\newblock In {\em ECCV}, 2022.

\bibitem{shi2023flowformer++}
Xiaoyu Shi, Zhaoyang Huang, Dasong Li, Manyuan Zhang, Ka~Chun Cheung, Simon
  See, Hongwei Qin, Jifeng Dai, and Hongsheng Li.
\newblock Flowformer++: Masked cost volume autoencoding for pretraining optical
  flow estimation.
\newblock In {\em CVPR}, 2023.

\bibitem{jonschkowski2020matters}
Rico Jonschkowski, Austin Stone, Jonathan~T Barron, Ariel Gordon, Kurt
  Konolige, and Anelia Angelova.
\newblock What matters in unsupervised optical flow.
\newblock In {\em ECCV}, 2020.

\bibitem{stone2021smurf}
Austin Stone, Daniel Maurer, Alper Ayvaci, Anelia Angelova, and Rico
  Jonschkowski.
\newblock Smurf: Self-teaching multi-frame unsupervised raft with full-image
  warping.
\newblock In {\em CVPR}, 2021.

\bibitem{janai2018unsupervised}
Joel Janai, Fatma Guney, Anurag Ranjan, Michael Black, and Andreas Geiger.
\newblock Unsupervised learning of multi-frame optical flow with occlusions.
\newblock In {\em ECCV}, 2018.

\bibitem{meister2018unflow}
Simon Meister, Junhwa Hur, and Stefan Roth.
\newblock Unflow: Unsupervised learning of optical flow with a bidirectional
  census loss.
\newblock In {\em AAAI}, 2018.

\bibitem{wang2018occlusion}
Yang Wang, Yi~Yang, Zhenheng Yang, Liang Zhao, Peng Wang, and Wei Xu.
\newblock Occlusion aware unsupervised learning of optical flow.
\newblock In {\em CVPR}, 2018.

\bibitem{luo2021upflow}
Kunming Luo, Chuan Wang, Shuaicheng Liu, Haoqiang Fan, Jue Wang, and Jian Sun.
\newblock Upflow: Upsampling pyramid for unsupervised optical flow learning.
\newblock In {\em CVPR}, 2021.

\bibitem{he2020momentum}
Kaiming He, Haoqi Fan, Yuxin Wu, Saining Xie, and Ross Girshick.
\newblock Momentum contrast for unsupervised visual representation learning.
\newblock In {\em CVPR}, 2020.

\bibitem{gidaris2018unsupervised}
Spyros Gidaris, Praveer Singh, and Nikos Komodakis.
\newblock Unsupervised representation learning by predicting image rotations.
\newblock In {\em ICLR}, 2018.

\bibitem{lai2017semi}
Wei-Sheng Lai, Jia-Bin Huang, and Ming-Hsuan Yang.
\newblock Semi-supervised learning for optical flow with generative adversarial
  networks.
\newblock In {\em NIPS}, 2017.

\bibitem{im2022semi}
Woobin Im, Sebin Lee, and Sung-Eui Yoon.
\newblock Semi-supervised learning of optical flow by flow supervisor.
\newblock In {\em ECCV}, 2022.

\bibitem{yan2020optical}
Wending Yan, Aashish Sharma, and Robby~T Tan.
\newblock Optical flow in dense foggy scenes using semi-supervised learning.
\newblock In {\em CVPR}, 2020.

\bibitem{he2016deep}
Kaiming He, Xiangyu Zhang, Shaoqing Ren, and Jian Sun.
\newblock Deep residual learning for image recognition.
\newblock In {\em CVPR}, 2016.

\bibitem{deng2009imagenet}
Jia Deng, Wei Dong, Richard Socher, Li-Jia Li, Kai Li, and Li~Fei-Fei.
\newblock Imagenet: A large-scale hierarchical image database.
\newblock In {\em CVPR}, 2009.

\bibitem{dong2023rethinking}
Qiaole Dong, Chenjie Cao, and Yanwei Fu.
\newblock Rethinking optical flow from geometric matching consistent
  perspective.
\newblock In {\em CVPR}, 2023.

\bibitem{pan2023drag}
Xingang Pan, Ayush Tewari, Thomas Leimk{\"u}hler, Lingjie Liu, Abhimitra Meka,
  and Christian Theobalt.
\newblock Drag your gan: Interactive point-based manipulation on the generative
  image manifold.
\newblock In {\em ACM SIGGRAPH}, 2023.

\bibitem{lu2020devon}
Yao Lu, Jack Valmadre, Heng Wang, Juho Kannala, Mehrtash Harandi, and Philip
  Torr.
\newblock Devon: Deformable volume network for learning optical flow.
\newblock In {\em WACV}, 2020.

\bibitem{jaderberg2015spatial}
Max Jaderberg, Karen Simonyan, Andrew Zisserman, et~al.
\newblock Spatial transformer networks.
\newblock In {\em NIPS}, 2015.

\bibitem{feichtenhofer2022masked}
Christoph Feichtenhofer, Yanghao Li, Kaiming He, et~al.
\newblock Masked autoencoders as spatiotemporal learners.
\newblock In {\em NIPS}, 2022.

\bibitem{johnson2016perceptual}
Justin Johnson, Alexandre Alahi, and Li~Fei-Fei.
\newblock Perceptual losses for real-time style transfer and super-resolution.
\newblock In {\em ECCV}, 2016.

\bibitem{zhao2022thin}
Jian Zhao and Hui Zhang.
\newblock Thin-plate spline motion model for image animation.
\newblock In {\em CVPR}, 2022.

\bibitem{xu2022gmflow}
Haofei Xu, Jing Zhang, Jianfei Cai, Hamid Rezatofighi, and Dacheng Tao.
\newblock Gmflow: Learning optical flow via global matching.
\newblock In {\em CVPR}, 2022.

\bibitem{Mayer_A_Large_Dataset}
Nikolaus Mayer, Eddy Ilg, Philip Häusser, Philipp Fischer, Daniel Cremers,
  Alexey Dosovitskiy, and Thomas Brox.
\newblock A large dataset to train convolutional networks for disparity,
  optical flow, and scene flow estimation.
\newblock In {\em CVPR}, 2016.

\bibitem{Butler_A_Naturalistic}
Daniel~J. Butler, Jonas Wulff, Garrett~B. Stanley, and Michael~J. Black.
\newblock A naturalistic open source movie for optical flow evaluation.
\newblock In {\em ECCV}, 2012.

\bibitem{Menze2015CVPR}
Moritz Menze and Andreas Geiger.
\newblock Object scene flow for autonomous vehicles.
\newblock In {\em CVPR}, 2015.

\bibitem{kondermann2016hci}
Daniel Kondermann, Rahul Nair, Katrin Honauer, Karsten Krispin, Jonas Andrulis,
  Alexander Brock, Burkhard Gussefeld, Mohsen Rahimimoghaddam, Sabine Hofmann,
  Claus Brenner, et~al.
\newblock The hci benchmark suite: Stereo and flow ground truth with
  uncertainties for urban autonomous driving.
\newblock In {\em CVPRW}, 2016.

\bibitem{janai2017slow}
Joel Janai, Fatma Guney, Jonas Wulff, Michael~J Black, and Andreas Geiger.
\newblock Slow flow: Exploiting high-speed cameras for accurate and diverse
  optical flow reference data.
\newblock In {\em CVPR}, 2017.

\bibitem{Caelles_davis_2019}
Sergi Caelles, Jordi Pont-Tuset, Federico Perazzi, Alberto Montes,
  Kevis-Kokitsi Maninis, and Luc {Van Gool}.
\newblock The 2019 davis challenge on vos: Unsupervised multi-object
  segmentation.
\newblock {\em arXiv:1905.00737}, 2019.

\bibitem{loshchilov2018decoupled}
Ilya Loshchilov and Frank Hutter.
\newblock Decoupled weight decay regularization.
\newblock In {\em ICLR}, 2018.

\bibitem{jiang2021learningtoestimate}
Shihao Jiang, Dylan Campbell, Yao Lu, Hongdong Li, and Richard Hartley.
\newblock Learning to estimate hidden motions with global motion aggregation.
\newblock In {\em ICCV}, 2021.

\bibitem{sun2021loftr}
Jiaming Sun, Zehong Shen, Yuang Wang, Hujun Bao, and Xiaowei Zhou.
\newblock Loftr: Detector-free local feature matching with transformers.
\newblock In {\em CVPR}, 2021.

\bibitem{tang2021quadtree}
Shitao Tang, Jiahui Zhang, Siyu Zhu, and Ping Tan.
\newblock Quadtree attention for vision transformers.
\newblock In {\em ICLR}, 2021.

\end{thebibliography}
}

% \clearpage
\setcounter{page}{1}
\maketitlesupplementary

%%%TODO 1118
% 1. inference pipeline figure and related statements
% 2. detail table of log graph in main paper
% 3. refine kitti test pic
% 4. change symbol in algorithm blocks, keep consistent with main paper

%%%
% 1. inference settings: duplicate with 6.
% 2. hyperparameter settings
% 3. more pics
% 4. detail log of figure 4 semi-supervised learning gains: origin is the full validation results
% 5. table of ablation K
% 6. pseudo-code for training and inference strategies
%%%

\section{Pipeline}
\subsection{Training}
%% algorithm 1: AIL
\begin{algorithm*}[!b]
    \caption{Apparent Information Learning}\label{alg:ail}
    \SetAlgoLined
    \KwData{Labeled Data $D_l$}
    \KwIn{$I_{t_0}, I_{t_1}, f_{<t_0,t_1>}, O_{<t_0,t_1>}$}
    \KwOut{$\hat{f}_{<t_0,t_1>}$}
    \ForEach{$I_{t_0}, I_{t_1}, f_{<t_0,t_1>}, O_{<t_0,t_1>}$ in $D_l$}{
        
        $I_{t_0}$, $I_{t_1}$ $\leftarrow$ Input frames at $t_0$, $t_1$\;
        $f_{<t_0, t_1>}$ $\leftarrow$ Ground truth optical flow from $I_{t_0}$ to $I_{t_1}$\;
        $O_{<t_0, t_1>}$ $\leftarrow$ Corresponding occlusion map\;
        $F_{t_0}$ $\leftarrow$ AIE($I_{t_0}$) \tcp*[r]{AIE: Apparent Information Encoder}
        $F_{t_1}$ $\leftarrow$ AIE($I_{t_1}$) \;
        $\hat{f}_{<t_0, t_1>}$ $\leftarrow$ MD($F_{t_0}$, $F_{t_1}$)  \tcp*[r]{MD: Motion Decoder}        
        $\hat{f}_{<t_1,t_0>}$ $\leftarrow$ MD($F_{t_1}, F_{t_0}$) \;
        $\hat{O}_{<t_0, t_1>}, \tilde{I}_{t_1}$ $\leftarrow$ WarpNet($\hat{f}_{<t_0,t_1>}, \hat{f}_{<t_1,t_0>}, I_{t_0})$ \;

        \SetKwFunction{Loss}{Loss}
        \SetKwProg{Fn}{Function}{:}{}
        \Fn{\Loss{$\hat{A}$, $A$}}{  
            loss $\leftarrow$ 0\;
            \For{each pixel $(x, y)$}{
                loss $\leftarrow$ loss + $\lvert \hat{A}(x, y) - A(x, y) \rvert$ \tcp*[r]{$\hat{A}$ is predicted. $A$ is the ground truth.}
            }
            \Return loss\;
        }

        \SetKwFunction{VGG}{VGG}
        \SetKwProg{Fn}{Function}{:}{}
        \Fn{\VGG{$I$}}{
            $List_{vgg_{map}}$ $\leftarrow$ [ ]\;
            $x$ $\leftarrow$ $I$ \;
            $N \leftarrow $ number of Vgg layers \;  
            \For{$i \gets 0, N-1$}{
                $x$ $\leftarrow$ $V_i(x)$ \tcp*[r]{$V_i$ is $i$-th layer in VGG.}
                Append $x$ to $List_{vgg_{map}}$ \;
            } 
            \Return $List_{vgg_{map}}$\;
        }
        $\mathbf{L}_{1}$ $\leftarrow$ \Loss{$\hat{f}_{<t_0, t_1>}$, $f_{<t_0, t_1>}$}\;
        $\mathbf{L}_{Occ} \leftarrow$ \Loss{$\hat{O}$, $O$} \;
        $\mathbf{L}_{Perceptual} \leftarrow \lvert$ \VGG{$\tilde{I}_{t_1}$} $-$ \VGG{$I_{t_1}$} $\rvert$\;
    }
\end{algorithm*}

Overall training pipeline is provided in \cref{alg:ail,alg:kgl}.
In the apparent information learning phase as \cref{alg:ail}, we enable the WarpNet with Sintel training stage. To obtain the occlusion map and the warped frame at the same time, we input $I_{t_0}$ and corresponding predicted bi-directional optical flow, $\hat{f}_{<t_0,t_1>}$ and $\hat{f}_{<t_1,t_0>}$. Here $\hat{f}_{<t_1, t_0>}$ is obtained by switching $F_{t_0}$ and $F_{t_1}$ in Motion Decoder. To adapt the number of channels, we build a convolutional layer to convert $I_{t_0}$ to feature domain and another convolutional layer to convert the warped frame back to image domain. In the kinetics guided learning phase as \cref{alg:kgl}, different from warping frame, to obtain the warped feature, we simply input the feature $F_{t_0}$ and corresponding optical flow. Notably, we duplicate the optical flow to fit the channels of WarpNet. Afterwards, we achieve warping $F_{t_0}$ to $F_{t_0+\Delta t}$.

%% algorithm 2: KGL
\begin{algorithm*}[htbp]
    \caption{Kinetics Guided Learning}\label{alg:kgl}
    \SetAlgoLined
    \KwData{Unlabeled Data $D_u$}
    \KwIn{$I_{t_0}, I_{t_1}$}
    \KwOut{$\hat{f}_{<t_0,t_1>}$ predicted optical flow from $t_0$ to $t_1$}
    \ForEach{$I_{t_0}, I_{t_1}$ in $D_u$}{
        $I_{t_0}$, $I_{t_1}$ $\leftarrow$ Input frames at $t_0$, $t_1$\;
        $F_{t_0}$ $\leftarrow$ AIE($I_{t_0}$) \tcp*[r]{AIE: Apparent Information Encoder}
        $F_{t_1}$ $\leftarrow$ AIE($I_{t_1}$) \;
        $\hat{f}_{<t_0, t_1>}$ $\leftarrow$ MD($F_{t_0}$, $F_{t_1}$)  \tcp*[r]{MD: Motion Decoder}

        % motion supervisor
        Random $\Delta t \in (0,1)$ \;
        \SetKwFunction{KineticsGuidedMotionGenerator}{KineticsGuidedMotionGenerator}
        \SetKwProg{Fn}{Function}{:}{}
        \Fn{\KineticsGuidedMotionGenerator{$f, \Delta t$}}{
            $f_{supervisor} \leftarrow f \times \Delta t$ \;
            \Return{$f_{supervisor}$} \;
        }
        $f_{<t_0, t_0+\Delta t>} \leftarrow $ \KineticsGuidedMotionGenerator{$\hat{f}_{<t_0,t_1>}, \Delta t$} \;
        $F_{t_0+\Delta t} \leftarrow $ WarpNet($F_{t_0}, f_{<t_0, t_0+\Delta t>}$) \;
        $\hat{f}_{<t_0, t_0+\Delta t>} \leftarrow$ MD($F_{t_0}, F_{t_0+\Delta t}$) \;
        $\mathbf{L}_{SSL} \leftarrow$ \Loss{$\hat{f}_{<t_0, t_0+\Delta t>}, f_{<t_0, t_0+\Delta t>}$} \;
    }
\end{algorithm*}

\subsection{Inference}
\begin{figure*}
    \includegraphics[width=1.0\linewidth]{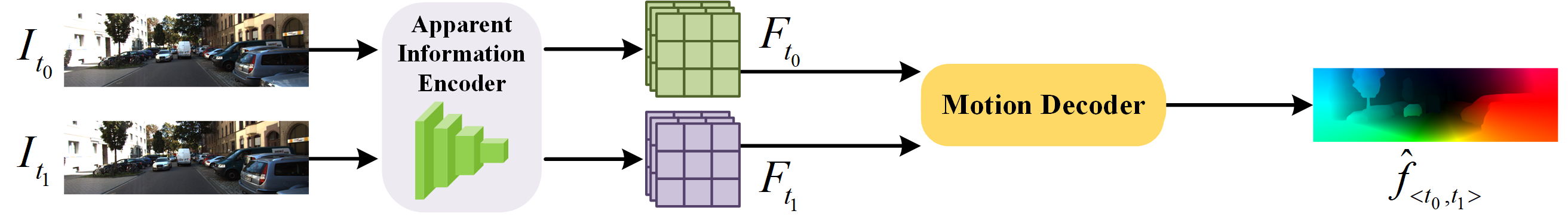}
    \caption{Inference pipeline. Apparent Information Encoder can encode the two frames $I_{t_0}$ and $I_{t_1}$ into features $F_{t_0}$ and $F_{t_1}$. With Motion Decoder, we obtain the corresponding optical flow $\hat{f}_{t_0, t_1}$ from $t_0$ towards $t_1$.\label{fig:supp_inference}}
\end{figure*}

The inference process is shown in \cref{fig:supp_inference}. We can obtain the optical flow from Apparent Information Encoder and Motion Decoder.

\section{More Experiments}
\label{sec:supp_exps}
To clarify the experimental results, some related experimental settings are reiterated.

\subsection{Implementation Details}
\textbf{Hyperparameter settings.} We declare the relevant hyperparameters, shown in \cref{tab:supp_hyperparameter}. Not only we use prevailing optical flow datasets, e.g., FlyingChairs(C)~\cite{Dosovitskiy_FlowNet}, FlyingThings3D(T)~\cite{Mayer_A_Large_Dataset}, MPI Sintel(S)~\cite{Butler_A_Naturalistic}, KITTI-2015(K)~\cite{Menze2015CVPR}, and HD1K(H)~\cite{kondermann2016hci}, but introduce some unlabeled datasets DAVIS(D)\cite{Caelles_davis_2019} and High Speed Sintel(HS)~\cite{janai2017slow}. \cref{tab:supp_hyperparameter} shows the used training datasets for each training stage.

% hyperparameters settings
\begin{table}[h]
    \caption{Hyperparameter settings.\label{tab:supp_hyperparameter}}  % 
    \resizebox{1.0\linewidth}{!}{
    \begin{tabular}{c|cccc c}
        \toprule
        stage & learning rate & image size & batch size & training steps & training dataset \\
        \midrule
        FlyingChairs       & 0.0004   & $384\times512$  & 16& 150k & C \\
        FlyingThings3D     & 0.0002   & $384\times768$  & 8 & 400k & T \\
        Sintel             & 0.0002   & $320\times896$  & 8 & 150k & T+S+K+H \\
        KITTI              & 0.0002   & $320\times1152$ & 8 & 150k & K \\
        Self-Supervised    & 0.000125 & $480\times832$  & 8 & 100k & D+HS \\
        \bottomrule
    \end{tabular}
    }
\end{table}

\subsection{More Comparisons}

\noindent\textbf{GPU memory usage.} \cref{tab:supp_gpu_memory} shows that although the number of parameters of our model is around double of RAFT~\cite{teed2020raft}, our method consumes the least computational time, only 55ms for a single pair of images in the FlyingChairs dataset. The GRU iteration number of RAFT~\cite{teed2020raft} and GMA\cite{jiang2021learningtoestimate} is set to 32 equally for comparison.
%GPU memory usage
\begin{table}[htbp]
    \centering
    \caption{Comparison of parameters, GPU memory, and time. This table is measured at FlyingChairs resolution 368$\times$496 with a batch size of 8. The results of the KITTI-2015 test set are provided on the leaderboard.\label{tab:supp_gpu_memory}}
    % \vspace{-1em}
    \resizebox{1.0\linewidth}{!}{
    \begin{tabular}{l|c c c c cc}
    \toprule
    \multirow{2}{*}{Method} & Params & Memory & Time & \multirow{2}{*}{$\Delta\frac{Memory(GB)}{Params(M)}$} & KITTI-15(test) \\ \cline{6-6}
    \multicolumn{1}{c|}{} & (M) & (GB) & (ms/pair) & \multicolumn{1}{c}{} & Fl-all \\
    \midrule
    RAFT\cite{teed2020raft}  & 5.26 & \textbf{10.57} & 155 & 2.01  & 5.10\\
    GMA\cite{jiang2021learningtoestimate} & 5.88 & 13.46 & 99 & 2.29  & 5.15\\
    GMFlow\cite{xu2022gmflow}  & \textbf{4.68} & 13.27 & 167 & 2.84  & 9.32 \\
    Ours & 11.81 & 16.93 & \textbf{55} & \textbf{1.43} & \textbf{4.94}\\
    \bottomrule
    \end{tabular}
    }
\end{table}

\noindent\textbf{Semi-supervised Gains.}
\cref{fig:fig4_ss_gains} and \cref{tab:ss_gains} show our proposed unsupervised learning can achieve lower EPE on the Sintel training set.
%% unsupervised learning graph
\begin{figure}[h]
    \includegraphics[width=1.0\linewidth]{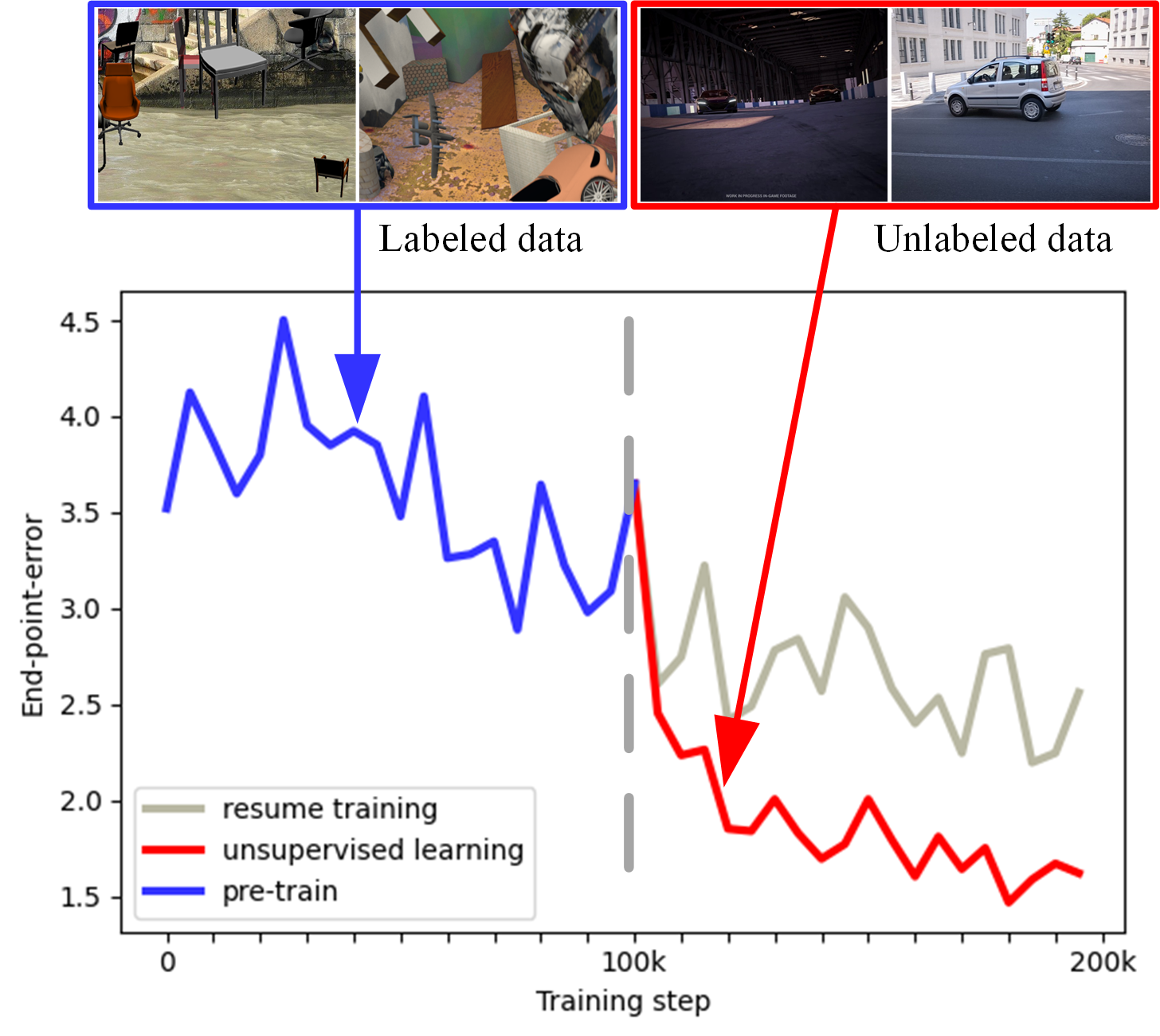}
    \caption{EPE on Sintel training set. Our model uses more unlabeled datasets DAVIS\cite{Caelles_davis_2019} and High Speed Sintel\cite{janai2017slow} to finetune our model. Our method decreases more loss compared to the supervised style with the same iterations.\label{fig:fig4_ss_gains}}
\end{figure}
%% unsupervised learning table

\begin{table}[htbp]
    \centering
    \caption{Our proposed unsupervised learning (Kinetics-guided Learning) strategy achieves lower EPE on Sintel training set with the same training steps. \label{tab:ss_gains}}
    % \vspace{-1em}
    \resizebox{1.0\linewidth}{!}{
    \begin{tabular}{l|cccccccccc}
    \toprule
    % \diagbox{Method}{EPE}{Training Step} & 20k & 40k & 60k & 80k & 100k & 120k & 140k & 160k & 180k & 200k \\
    Strategy & 20k & 40k & 60k & 80k & 100k & 120k & 140k & 160k & 180k & 200k \\
    \midrule
    Supervised      & 3.800 & 3.924 & 3.262 & 3.645 & 3.651 & - & - & - & - & -  \\
    Resume training & - & - & - & - & - & 3.223 & 2.841 & 2.585 & 2.762 & 2.562 \\
    Unsupervised    & - & - & - & - & - & 2.064 & 1.629 & 1.596 & 1.552 & 1.421 \\    
    \bottomrule
    \end{tabular}
    }
\end{table}

%% tab: eval on train+test set

\begin{figure*}[t] 
    \centering
    % \vspace{-1em}
    \includegraphics[width=1.0\linewidth]{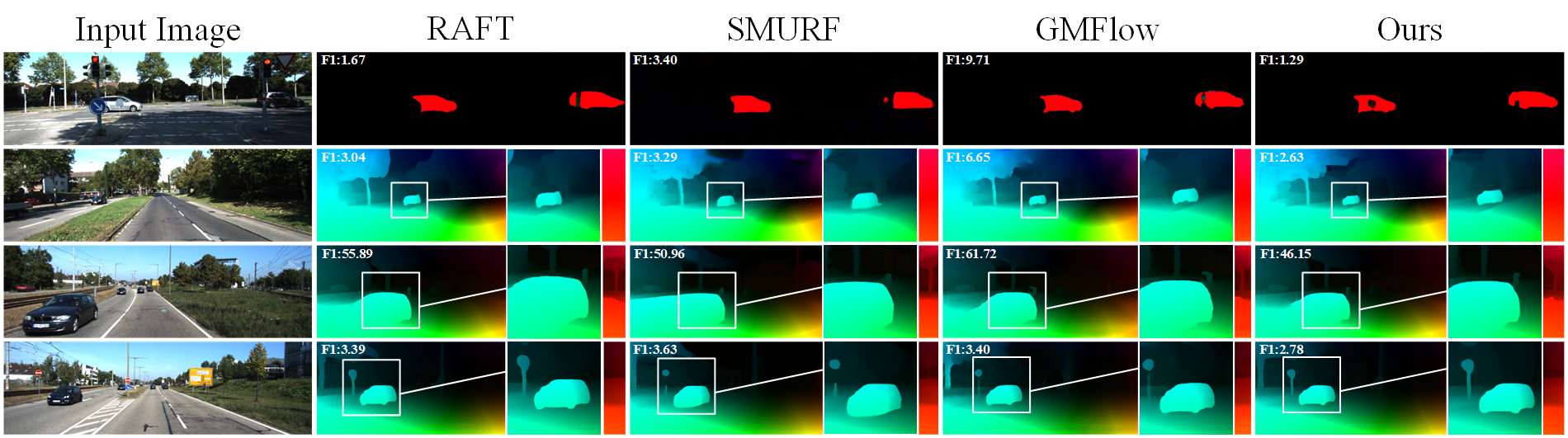}
    \caption{visual comparisons on KITTI test set with magnification. Obviously, our method can achieve more clearer object edges.\label{fig:supp_vis_kitti_mag}}
\end{figure*}

\begin{figure*}[t]
    \includegraphics[width=1.0\linewidth]{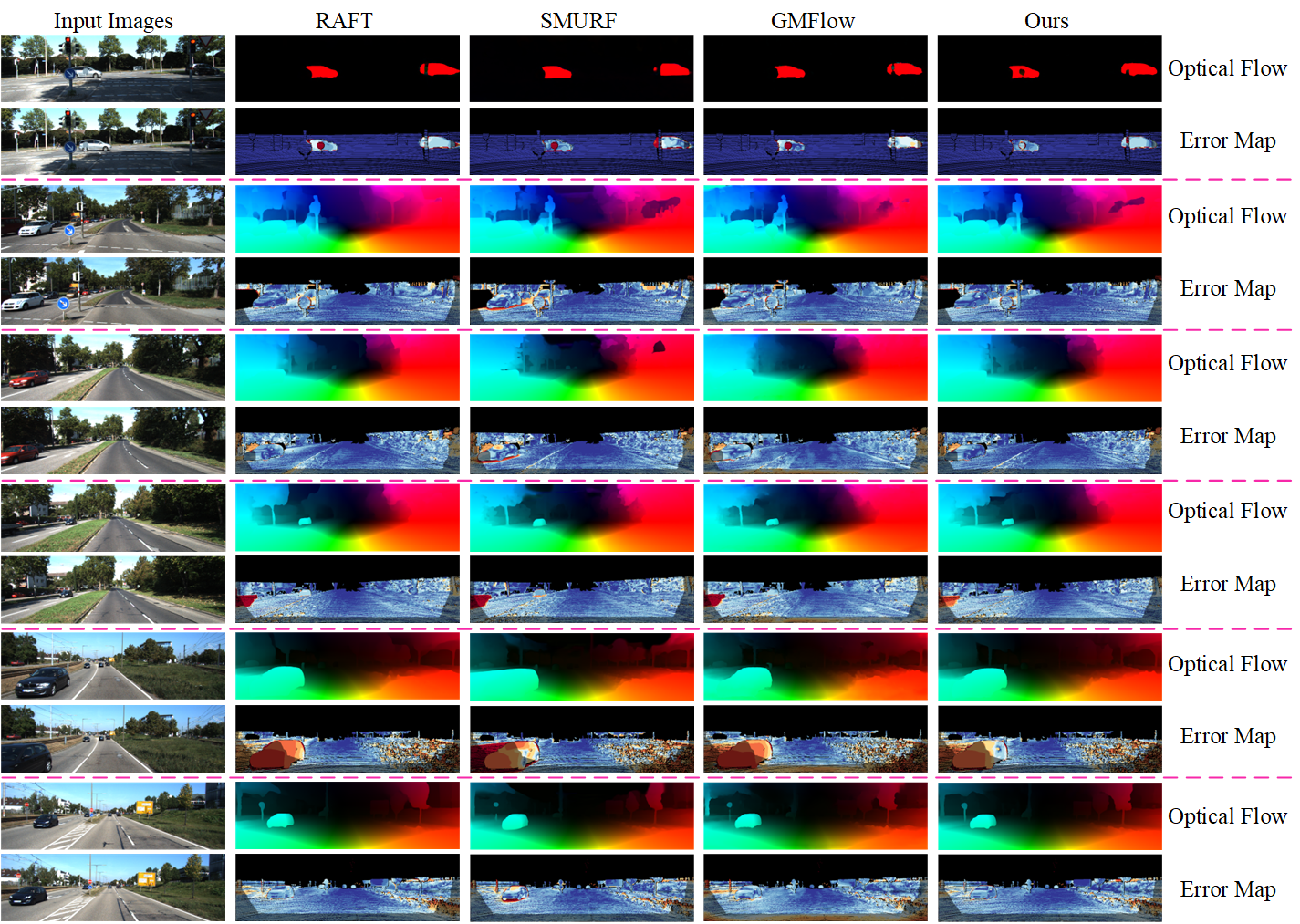}
    \caption{More visual comparisons on KITTI test set (better in color). For every two rows, the top is optical flow and the bottom is the corresponding error map where the more precise areas are in blue.\label{fig:supp_all_kitti_test}}
\end{figure*}

\noindent\textbf{Comparison with SOTA.} \cref{tab:supp_eval_testset} are quantitative results on the Sintel training set and KITTI-2015 benchmark. Supervised models are pre-trained on FlyingChairs and FlyingThings3D, then finetuned on Sintel and KITTI-2015. Some models also use HD1K dataset to finetune Sintel.
Without specific notation, the ablation studies in the main paper are performed with models pre-trained on FlyingChairs and FlyingThings3D.
\noindent\textbf{More detailed comparisons on KITTI-2015 benchmark.} \cref{tab:supp_index_comparison_of_all_noc} analyzes the ability of methods to estimate optical flow over the foreground, background, and all regions on KITTI-2015 test set.
% \cref{tab:supp_index_comparison_of_all_noc} provides quantitative comparisons of specific test images in KITTI-2015 benchmark.
\cref{fig:supp_all_kitti_test,fig:supp_vis_kitti_mag} show more visual performance on KITTI-2015 test set.
%overall fg bg all
% \begin{table}[h]
%     \caption{Comparison on KITTI-2015 test set from online leaderboard. `All' denotes evaluated on all pixels with ground truth and `Noc' denotes non-occluded pixels only. Fl-bg and Fl-fg evaluate the percentage of outliers averaged only over only background and foreground regions, respectively. The best results are \textbf{bolded}.}  % C+T
%     \resizebox{1.0\linewidth}{!}{
%     \begin{tabular}{l| c ccc c ccc}
%         \toprule
%         \multirow{2}{*}{Method} && \multicolumn{3}{c}{All} && \multicolumn{3}{c}{Noc} \\ \cline{3-5} \cline{7-9}
%         \multicolumn{1}{l|}{} && Fl-bg & Fl-fg & Fl-all && Fl-bg & Fl-fg & Fl-all \\ 
%         \midrule
%         RAFT\cite{teed2020raft}               && 4.74 & 6.87 & 5.10 && \textbf{2.87} & 3.98 & 3.07 \\
%         SMURF\cite{stone2021smurf}            && 4.74 & 6.87 & 5.10 && \textbf{2.87} & 3.98 & 3.07 \\
%         GMFlow\cite{xu2022gmflow}             && 9.67 & 7.57 & 9.32 && 3.65 & 4.46 & 3.80 \\
%         % UFlow\cite{jonschkowski2020matters}   && 9.78 & 17.87& 11.13 && 7.01 & 14.75 & 8.41 \\        
%         % SemARFlow\cite{Yuan_2023_ICCV_SemARFlow} && 7.48 & 12.91 & 8.38 && 4.58 & 9.30 & 5.43 \\
%         Ours                                  && \textbf{4.62} & \textbf{6.54} & \textbf{4.94} && \textbf{2.87} & \textbf{3.70} & \textbf{3.02}\\
%         \bottomrule
%     \end{tabular}
%     }
%     \label{tab:supp_comparison_detail_kitti}
% \end{table}

%overall+specific fg bg all
\begin{table}[h]
    \caption{Comparison on KITTI-2015 test set from online leaderboard. `All' denotes evaluated on all pixels with ground truth and `Noc' denotes non-occluded pixels only. Fl-bg and Fl-fg evaluate the percentage of outliers averaged only over only background and foreground regions, respectively.\label{tab:supp_index_comparison_of_all_noc}}  % C+T
    \resizebox{1.0\linewidth}{!}{
    \begin{tabular}{c l| c ccc c ccc}
        \toprule
        \multirow{2}{*}{Test Sequence} & \multirow{2}{*}{Method} && \multicolumn{3}{c}{All} && \multicolumn{3}{c}{Noc} \\ \cline{4-6} \cline{8-10}        
        \multicolumn{1}{c}{}& \multicolumn{1}{l|}{} && Fl-bg & Fl-fg & Fl-all && Fl-bg & Fl-fg & Fl-all \\ 
        \midrule
        \multirow{4}{*}{Overall} & RAFT\cite{teed2020raft} && 4.74 & 6.87 & 5.10 && \textbf{2.87} & 3.98 & 3.07 \\
        \multicolumn{1}{c}{} & SMURF\cite{stone2021smurf}  && 4.74 & 6.87 & 5.10 && \textbf{2.87} & 3.98 & 3.07 \\
        \multicolumn{1}{c}{} & GMFlow\cite{xu2022gmflow}   && 9.67 & 7.57 & 9.32 && 3.65 & 4.46 & 3.80 \\
        \multicolumn{1}{c}{} & Ours   && \textbf{4.62} & \textbf{6.54} & \textbf{4.94} && \textbf{2.87} & \textbf{3.70} & \textbf{3.02}\\
        \midrule
        \multirow{4}{*}{1}& RAFT\cite{teed2020raft}          && \underline{1.64} & \textbf{8.21} & \underline{2.37} && \underline{1.65} & \textbf{8.21} & \underline{2.46} \\
        \multicolumn{1}{c}{}& SMURF\cite{stone2021smurf}     && 2.71 & 13.85 & 3.95 && 2.58 & 13.85 & 3.97\\
        \multicolumn{1}{c}{}& GMFlow\cite{xu2022gmflow}      && 2.95 & 13.93 & 4.17 && 1.71 & 13.93 & 3.22\\
        \multicolumn{1}{c}{}& Ours                           && \textbf{1.45} & \underline{8.97} & \textbf{2.29} && \textbf{1.42} & \underline{8.97} & \textbf{2.35}\\
        \midrule
        % \multirow{4}{*}{3}& RAFT\cite{teed2020raft}          && 8.82 & 20.00 & 9.85 && 6.02 & 8.54 & 6.24 \\
        % \multicolumn{1}{c}{}& SMURF\cite{stone2021smurf}     && 8.55 & 6.98 & 8.40 && 6.53 & 7.59 & 6.02 \\
        % \multicolumn{1}{c}{}& GMFlow\cite{xu2022gmflow}      && 11.38 & 4.71 & 10.76 && 4.61 & 3.87 & 4.55 \\
        % % \multicolumn{1}{c}{}& MatchFlow                    && 8.23 & 42.14 & 11.37 && 5.69 & 27.58 & 7.57 \\
        % \multicolumn{1}{c}{}& Ours                           && 6.82 & 11.04 & 7.21 && 4.96 & 9.61 & 5.36\\
        % \midrule
        % \multirow{4}{*}{4}  & RAFT\cite{teed2020raft}        && 2.41 & 2.15 & 2.36 && 2.27 & 1.76 & 2.18 \\
        % \multicolumn{1}{c}{}& SMURF\cite{stone2021smurf}     && 5.81 & 16.79 & 7.63 && 6.20 & 15.99 & 8.03 \\        
        % \multicolumn{1}{c}{}& GMFlow\cite{xu2022gmflow}      && 10.85 & 14.46 & 11.45 && 4.92 & 13.98 & 6.61 \\
        % % \multicolumn{1}{c}{}& MatchFlow                    && 3.19 & 10.54 & 4.41 && 2.46 & 8.44 & 3.57 \\
        % \multicolumn{1}{c}{}& Ours                           && 3.88 & 10.07 & 4.91 && 3.16 & 8.57 & 4.17 \\
        % \midrule
        \multirow{4}{*}{7}  & RAFT\cite{teed2020raft}        && 1.34 & \underline{3.00} & \underline{1.67} && 1.34 & \underline{3.01} & \underline{1.65} \\
        \multicolumn{1}{c}{}& SMURF\cite{stone2021smurf}     && 1.51 & 11.18 & 3.40 && 1.51 & 11.36 & 3.35 \\        
        \multicolumn{1}{c}{}& GMFlow\cite{xu2022gmflow}      && \underline{1.14} & 9.71 & 2.82 && \underline{1.14} & 7.20 & 2.27 \\
        \multicolumn{1}{c}{}& Ours                           && \textbf{0.93} & \textbf{2.78} & \textbf{1.29} && \textbf{0.93} & \textbf{2.77} & \textbf{1.27}\\
        \midrule
        \multirow{4}{*}{11}  & RAFT\cite{teed2020raft}       && \underline{1.95} & \textbf{0.92} & \underline{1.77} && 1.98 & \underline{0.92} & 1.77 \\
        \multicolumn{1}{c}{}& SMURF\cite{stone2021smurf}     && 4.58 & 3.43 & 4.37 && 5.04 & 3.43 & 4.72 \\        
        \multicolumn{1}{c}{}& GMFlow\cite{xu2022gmflow}      && 1.97 & 1.25 & 1.84 && \underline{1.52} & 1.22 & \underline{1.46} \\
        \multicolumn{1}{c}{}& Ours                           && \textbf{1.42} & \underline{0.96} & \textbf{1.34} && \textbf{1.43} & \textbf{0.90} & \textbf{1.33}\\
        \midrule
        
        \multirow{4}{*}{13} & RAFT\cite{teed2020raft}        && \underline{3.54} & \underline{1.24} & \underline{3.26} && 1.61 & \textbf{0.93} & 1.53 \\
        \multicolumn{1}{c}{}& SMURF\cite{stone2021smurf}     && 4.24 & 8.05 & 4.71 && 2.23 & 9.14 & 3.03\\        
        \multicolumn{1}{c}{}& GMFlow\cite{xu2022gmflow}      && 7.82 & 1.57 & 7.06 && \textbf{1.31} & 1.67 & \underline{1.36} \\
        \multicolumn{1}{c}{}& Ours                           && \textbf{3.32} & \textbf{1.06} & \textbf{3.04} && \underline{1.33} & \underline{1.04} & \textbf{1.30}\\
        \midrule
        
        % \multirow{4}{*}{9}  & RAFT\cite{teed2020raft}        && 0.50 & 2.28 & 0.96 && 0.50 & 2.28 & 0.96 \\
        % \multicolumn{1}{c}{}& SMURF\cite{stone2021smurf}     && 1.57 & 7.65 & 3.14 && 1.57 & 7.65 & 3.14 \\        
        % \multicolumn{1}{c}{}& GMFlow\cite{xu2022gmflow}      && 0.48 & 2.90 & 1.10 && 0.48 & 2.90 & 1.10 \\
        % % \multicolumn{1}{c}{}& MatchFlow                    && 0.34 & 3.07 & 1.03 && 0.34 & 3.07 & 1.03 \\
        % \multicolumn{1}{c}{}& Ours                           && 0.51 & 2.30 & 0.97 && 0.51 & 2.30 & 0.97\\
        % \midrule
        \multirow{4}{*}{18} & RAFT\cite{teed2020raft}        && \underline{16.87} & 99.00 & 55.89 && 12.03 & 96.80 & 35.41 \\
        \multicolumn{1}{c}{}& SMURF\cite{stone2021smurf}     && 20.06 & \underline{85.09} & \underline{50.96} && 13.20 & 67.24 & 28.10 \\        
        \multicolumn{1}{c}{}& GMFlow\cite{xu2022gmflow}      && 28.16 & 98.79 & 61.72 && \textbf{9.87} & 96.14 & 33.67 \\
        % \multicolumn{1}{c}{}& MatchFlow             && 15.03 & 84.30 & 47.94 && 10.27 & 51.39 & 21.61 \\
        \multicolumn{1}{c}{}& Ours                           && \textbf{16.49} & \textbf{78.91} & \textbf{46.15} && \underline{11.13} & \textbf{39.21} & \textbf{18.87}\\
        \bottomrule
    \end{tabular}
    }
\end{table}

%%% ablations: k matches
\begin{table}[htbp]
    \centering
    \caption{Ablation study on $K$ trained on FlyingChairs for 150K iterations. The results are evaluated on a single V100 GPU when training on the FlyingChairs with a batch size of 16. These models are evaluated on the validation set of FlyingChairs. $\diamond$ represents the default settings.\label{tab:ablation_studies_knn}}
    \resizebox{1.0\linewidth}{!}{
    \begin{tabular}{l|c cccc c c c}
        \toprule
        \multirow{2}{*}{\textit{K}} && \multicolumn{4}{c}{FlyingChairs(val)} && Params & Training Time \\ \cline{3-6}
        \multicolumn{1}{l|}{}  && EPE & 1px & 3px & 5px &&\multicolumn{1}{c}{(M)} & \multicolumn{1}{c}{(ms/pair)}\\ 
        \midrule
        4    && 1.255 & 0.175 & 0.066 & 0.042 && 10.40 & 53.33 \\
        8    && 1.166 & 0.162 & 0.060 & 0.039 && 10.49 & 52.77 \\
        16   && 1.173 & 0.167 & 0.062 & 0.039 && 10.66 & 53.96 \\
        32   && 1.124 & 0.159 & 0.058 & 0.037 && 11.03 & 54.86 \\
        64\textsuperscript{$\diamond$}&& 1.096& 0.157 & 0.056 & 0.036 && 11.81 & 55.35 \\
        128  && 1.062 & 0.148 & 0.053 & 0.033 && 13.60 & 61.33 \\
        \bottomrule
    \end{tabular}
    }
    \vspace{-1em}
\end{table}

% eval on sintel and kitti test
\begin{table*}[t]
    \centering
    \caption{Quantitative results on the Sintel training set and KITTI-2015 benchmark. Supervised models are pre-trained on FlyingChairs(C) and FlyingThings3D(T), then finetuned on Sintel and KITTI-2015. Some models are also finetuned on the HD1K dataset. \textsuperscript{*} represents the results of warm-start strategy used in RAFT\cite{teed2020raft}. $\dagger$ denotes the unsupervised methods that are trained using images from the target domain. $\ddagger$ represents the semi-supervised learning framework. - indicates unavailable data. The Gain/Param. measures the parameter gains of Fl-all on KITTI-2015 test set and $\downarrow$ indicates lower is more efficient. The best results are \textbf{bolded} and the second best results are \underline{underlined}.\label{tab:supp_eval_testset}}
    \resizebox{1.0\linewidth}{!}{
    \begin{tabular}{l |c cc c cc c c c cc}
    % \hline
    \toprule
    \multirow{2}{*}{Method} && \multicolumn{2}{c}{Sintel(train)} && \multicolumn{2}{c}{KITTI-15(train)} && KITTI-15(test) && Params. & \multirow{2}{*}{Gain/Param.($\downarrow$)} \\ \cline{3-4} \cline{6-7} \cline{9-9} 
    \multicolumn{1}{c|}{}   && Clean & Final && EPE & Fl-all && Fl-all && (M) & \multicolumn{1}{c}{}\\ 
    \midrule
    UpFlow\cite{luo2021upflow}\textsuperscript{$\dagger$} && 2.33 & 2.67 && 2.45 & - && 9.38 && 3.5 & 2.69 \\
    Lai et al.\cite{lai2017semi}\textsuperscript{$\ddagger$} && 2.41 & 3.16 && 14.69 & 30.30 && 31.01 && - & - \\
    RAFT\cite{teed2020raft} && 0.76/0.77\textsuperscript{*} & 1.22/1.27\textsuperscript{*} && 0.63 & 1.50 && 5.10 && 5.3 & 0.97 \\
    SCV\cite{jiang2021learning_optical_flow}&& 0.79/0.86\textsuperscript{*} & 1.70/1.75\textsuperscript{*} && 0.75 & 2.10 && 6.17 && 5.3 & 1.16 \\
    GMFlow\cite{xu2022gmflow}               && - & - && - & - && 9.32 && 4.7 & 1.99\\
    MatchFlow(G)\cite{dong2023rethinking}   && \textbf{0.49} & \textbf{0.78} && \textbf{0.55} & \textbf{1.10} && \textbf{4.63} && 15.4 & 0.30\\
    Ours                                    && \underline{0.50} & \underline{0.87} && \textbf{0.55} & \underline{1.23} && \underline{4.94}   && 11.8 & 0.42 \\
    \bottomrule
    \end{tabular}
    }    
\end{table*}

% 2. KITTI test set with magnification areas

% 1. kitti test set

\subsection{Ablations}
\textbf{Value of $\mathbf{K}$.} \cref{tab:ablation_studies_knn} ablates that $K=64$ achieves a better speed-accuracy tradeoff.

%Further details are provided in supplementary materials.

% % 
% Having the supplementary compiled together with the main paper means that:
% % 
% \begin{itemize}
% \item The supplementary can back-reference sections of the main paper, for example, we can refer to \cref{sec:intro};
% \item The main paper can forward reference sub-sections within the supplementary explicitly (e.g. referring to a particular experiment); 
% \item When submitted to arXiv, the supplementary will already included at the end of the paper.
% \end{itemize}
% % 

% To split the supplementary pages from the main paper, you can use \href{https://support.apple.com/en-ca/guide/preview/prvw11793/mac#:~:text=Delete%20a%20page%20from%20a,or%20choose%20Edit%20%3E%20Delete).}{Preview (on macOS)}, \href{https://www.adobe.com/acrobat/how-to/delete-pages-from-pdf.html#:~:text=Choose%20%E2%80%9CTools%E2%80%9D%20%3E%20%E2%80%9COrganize,or%20pages%20from%20the%20file.}{Adobe Acrobat} (on all OSs), as well as \href{https://superuser.com/questions/517986/is-it-possible-to-delete-some-pages-of-a-pdf-document}{command line tools}.

\end{document}